%% file: main.tex
\documentclass{article} %
\usepackage{iclr2023_conference_arxiv,times}

\input{math_commands.tex}

\usepackage{microtype}
\usepackage{graphicx}
\usepackage{caption} 
\usepackage{subcaption} 
\usepackage{booktabs} 
\usepackage{xcolor}
\definecolor{customcolor}{HTML}{1456a6}
\usepackage[colorlinks,allcolors=customcolor]{hyperref}
\usepackage{tabularx}
\usepackage{multirow}
\usepackage{algorithm}
\usepackage{algorithmic}
\usepackage{makecell}
\usepackage{verbatim}
\usepackage{amsmath}
\usepackage{amsfonts}
\usepackage{amssymb}
\usepackage{mathtools}
\usepackage{amsthm}
\theoremstyle{plain}

\theoremstyle{definition}

\theoremstyle{remark}

\newcommand{\CC}{\mathcal{C}}

\usepackage[capitalize,noabbrev]{cleveref}
\usepackage{wrapfig}

\newcommand{\var}{\text{var}}

\title{Discovering Latent Knowledge in Language Models Without Supervision}

\iclrfinalcopy

\author{Collin Burns$^*$ \\ UC Berkeley \\
\And
Haotian Ye$^*$ \\ Peking University \\
\And
Dan Klein \\ UC Berkeley \\
\And
Jacob Steinhardt \\ UC Berkeley
}

\begin{document}
\maketitle
\def\thefootnote{*}\footnotetext{Equal contribution.}\def\thefootnote{\arabic{footnote}}
\begin{abstract}
Existing techniques for training language models can be misaligned with the truth:
if we train models with imitation learning, they may reproduce errors that humans make; if we train them to generate text that humans rate highly, they may output errors that human evaluators can't detect.
We propose circumventing this issue by directly finding latent knowledge inside the internal activations of a language model in a purely unsupervised way.
Specifically, we introduce a method for accurately answering yes-no questions given only unlabeled model activations.
It works by finding a direction in activation space that satisfies logical consistency properties, such as that a statement and its negation have opposite truth values.
We show that despite using no supervision and no model outputs, our method can recover diverse knowledge represented in large language models: across 6 models and 10 question-answering datasets, it outperforms zero-shot accuracy by 4\% on average.
We also find that it cuts prompt sensitivity in half and continues to maintain high accuracy even when models are prompted to generate incorrect answers.
Our results provide an initial step toward discovering what language models know, distinct from what they say, even when we don't have access to explicit ground truth labels.
\looseness=-1

\end{abstract}

\input{sections/1-intro}

\input{sections/2-method}

\input{sections/3-experiments}

\input{sections/4-related}

\input{sections/5-conclusion}

\section*{Acknowledgements}

We are very grateful to Jared Kaplan for helpful experiment suggestions and resources early on in the project. We thank Beth Barnes and Paul Christiano for valuable discussions regarding the longer-term impacts of this work. We are also grateful to Jessy Lin, Alex Pan, Ruiqi Zhong, Yaodong Yu, the anonymous reviewers, and several others for useful feedback on earlier versions of this paper. CB is supported by an Open Philanthropy AI Fellowship.

\bibliography{main}
\bibliographystyle{iclr2023_conference}

\appendix

\input{sections/6-appendix}

\end{document}

%% file: math_commands.tex
\usepackage{amsmath,amsfonts,bm}

\def\eqref#1{equation~\ref{#1}}

\def\1{\bm{1}}

\DeclareMathAlphabet{\mathsfit}{\encodingdefault}{\sfdefault}{m}{sl}
\SetMathAlphabet{\mathsfit}{bold}{\encodingdefault}{\sfdefault}{bx}{n}

%% file: sections/1-intro.tex
\section{Introduction}
\label{sec:intro}

The increasing deployment of language models in real-world applications opens up exciting possibilities, but it also raises the stakes of AI research and presents new risks \citep{Bommasani2021OnTO, Weidinger2021EthicalAS, Bender2021OnTD}.
One of these risks is that language models do not always output text that is true \citep{Evans2021TruthfulAD,Hendrycks2021UnsolvedPI,Kenton2021AlignmentOL}.

Common training objectives can cause models to learn internal representations related to truth, since truth is a useful feature for many tasks.
However, these objectives can also cause language models to output text that is false, at least in some circumstances.
For example, if we train a model to imitate human-generated text, it may learn to output common misconceptions \citep{Lin2022TruthfulQAMH}.
Or if we train a chat bot to optimize a reward such as engagement, it may learn to generate text that is compelling but false \citep{Roller2021RecipesFB}.
If we try to reward model outputs that look true, a model may still learn to output false text if human raters can't evaluate the correctness of that text \citep{Kenton2021AlignmentOL}.

In each case, this is an issue that stems from the misalignment between a training objective and the truth.
As models are applied to more complex domains, human supervision may become less effective at mitigating this misalignment.
Moreover, because this is a problem with the training objective rather than a model's capabilities, it likely won't be solved by scaling up models alone.

\begin{figure}[t]
    \centering
    \includegraphics[width=\textwidth]{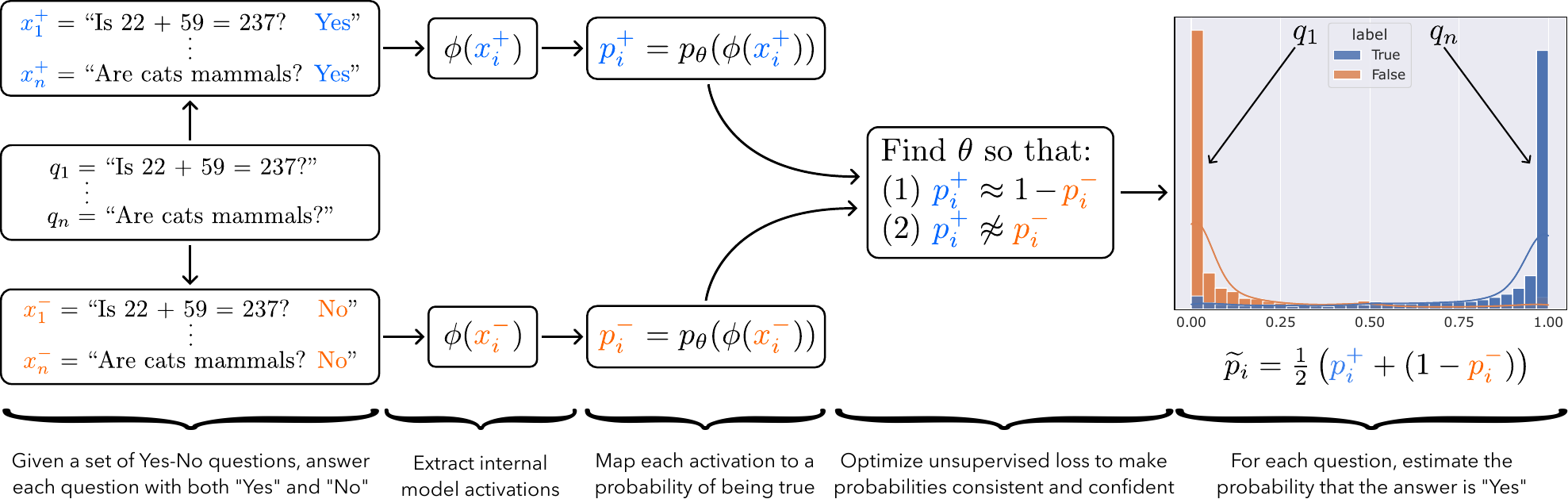}
    \caption{An illustration of our method, Contrast-Consistent Search (CCS). 	
    For each yes-no question $q_i$, we let $x_i^+$ and $x_i^-$ be the natural language statements where we answer $q_i$ as ``Yes'' and ``No'' respectively.
    Answering the question $q_i$ then amounts to determining which of $x_i^+$ or $x_i^-$ is true.
    We compute probabilities $p^+_i$ and $p^-_i$ that $x_i^+$ and $x_i^-$ are true respectively using a learned mapping from the hidden states to a number between $0$ and $1$. 
    We search for a mapping such that that the probabilities are both confident and consistent.
    On the right, we show a histogram of the ``Yes'' probabilities, $\tilde{p}_i = 0.5\cdot(p_i^+ + (1-p_i^-))$, learned by our method on the unlabeled train split of the COPA dataset \citep{roemmele2011choice} with the UnifiedQA model \citep{Khashabi2020UnifiedQACF}. 
    Our method uses no labels and no model outputs, but still learns to accurately answers questions.
    }
    \label{fig:pipeline}
    \vspace{-10pt}
\end{figure}

We propose a different approach for addressing this misalignment: using models to answer questions in a purely \textit{unsupervised} way. 
Intuitively, instead of trying to explicitly, externally specify truth, we search for implicit, internal ``beliefs'' or ``knowledge'' learned by a model.
We approach this problem by leveraging the fact that a model's representation of truth must satisfy logical consistency properties, which are unlikely to be satisfied by many other features.

We implement this idea by introducing Contrast-Consistent Search (CCS), a method that learns a linear projection of the hidden states that is consistent across negations, as illustrated in Figure~\ref{fig:pipeline}.
We find that despite its simplicity and despite not having access to any labels or model outputs, CCS can accurately recover knowledge from model representations: evaluated across 6 models and 10 question-answering datasets, CCS outperforms the accuracy of strong zero-shot baselines by 4\% on average (\Cref{subsec:ccs_outperforms}). 
The resulting classifier is also less sensitive to different prompts than zero-shot, cutting the standard deviation in accuracy in half.
Additionally, we try deliberately prompting models to make incorrect outputs, which should intuitively change what models say but which shouldn't affect their latent knowledge. 
We find that this causes zero-shot accuracy to drop by up to 9.5\% (\Cref{subsec:ccs_robust}) without decreasing the accuracy of CCS.

We systematically analyze CCS to understand the features it discovers.
We show that it transfers across unrelated tasks, suggesting that models may have a task-agnostic representation of the truth and that CCS is able to approximately discover it (\Cref{subsec:transfer}). 
Moreover, CCS sometimes works best using the hidden states in the \textit{middle} layers of a network and can work even when model outputs aren't very informative, suggesting that it can leverage different features from those used by the outputs (\Cref{subsec:not_model_outputs}).
Finally, we show that representations of truth tend to be salient in models: they can often be found without much data, and they can often be found by taking the top principal component of a slightly modified representation space (\Cref{sec:CRC}).

Most existing techniques for making models truthful use human supervision to explicitly specify what is correct.
However, it is not feasible to provide supervision in some settings.
Our work serves as a proof of concept that an external source of ground truth may not actually be necessary: we may instead be able to find a model's latent representation of truth, independent of what a model says, without using any supervision in the first place.

%% file: sections/2-method.tex
\vspace{-5pt}
\section{Problem Statement and Framework}
\label{sec:method}
\vspace{-3pt}

In this section we describe our problem setup in more detail and introduce Contrast-Consistent Search (CCS), a method for discovering latent knowledge in language models without supervision.

\vspace{-5pt}
\subsection{Problem: Discovering Latent Knowledge}
\label{subsec:problem_statement}
\vspace{-3pt}
Given a pre-trained neural language model and a set $q_1, \ldots, q_n$ of yes-no questions\footnote{
Technically, we only require that there are two mutually exclusive answers. 
For example, we can also use the labels ``positive'' and ``negative'' for sentiment classification.
Moreover, our setup can easily extend to the case where we want to evaluate the truth of a set of statements instead of answering a set of questions.
}, our goal is to answer each $q_i$ correctly.
Here, $q_i$ can be any question with a well-defined answer, including procedural questions like ``Is 22+59 = 237?'', for which the answer is ``No'', and factual questions like ``Are cats mammals?'', for which the answer is ``Yes''.

Importantly, we want methods that do not rely on the model generating correct outputs and that do not rely on external supervision.
Instead, we turn to the model's unlabeled hidden representations. 
Specifically, let $\phi(x) \in \mathbb{R}^d$ denote some feature representation on a natural langauge input $x$, such as the hidden states of a Transformer-based language model. 
Our goal is to answer the questions $q_1, \ldots, q_n$ only given access to $\phi(\cdot)$. 
In \Cref{subsec:method_ccs} we introduce a method for this problem that attains high accuracy (\Cref{sec:experiments}), demonstrating that this task is tractable.

\vspace{-5pt}
\subsection{Method: Contrast-Consistent Search}
\label{subsec:method_ccs}
\vspace{-3pt}
To make progress on the goal described above, we exploit the fact that truth has special structure: it satisfies consistency properties that few other features in a language model are likely to satisfy.
Our method, Contrast-Consistent Search (CCS), leverages this idea by finding a direction in activation space that is consistent across negations.
As we illustrate in \Cref{fig:pipeline}, CCS works by (1) answering each question $q_i$ as both ``Yes'' ($x_i^+$) and ``No'' ($x_i^-$), (2) computing the representations $\phi(x_i^+)$ and $\phi(x_i^-)$ of each answer, (3) mapping the answer representations to probabilities $p_i^+$ and $p_i^-$ of being true, then (4) optimizing that mapping so that the probabilities are both consistent and confident.

Concretely, the input to CCS is a set of Yes-No questions, $q_1, \ldots, q_n$, and access to a pretrained model's representations, $\phi(\cdot)$; the output of CCS is a lightweight probe on top of $\phi(\cdot)$ that can answer new questions.
Here, $\phi(\cdot)$ is fixed but should contain useful information about the answers to $q_1, \ldots, q_n$, in the sense that if one \textit{did} (hypothetically) have access to the ground-truth labels for $q_1, \ldots, q_n$, one would be able to train a small supervised probe on $\phi(\cdot)$ that attains high accuracy.
Importantly, CCS does not modify the weights of the pretrained model and it does not use labels.

\textbf{Constructing contrast pairs.}
An important property that truth satisfies is negation consistency: the answer to a clear-cut question cannot be both ``Yes'' and ``No'' at the same time, as these are negations of each other.
Probabilistically, for each question $q_i$, the probability that the answer to $q_i$ is ``Yes'' should be one minus the probability that the answer to $q_i$ is ``No''.
To use this property, we begin by constructing contrast pairs: for each question $q_i$, we answer $q_i$ both as ``Yes'', resulting in the new natural language statement $x_i^+$, and as ``No'', resulting in the natural language statement $x_i^-$.
We illustrate this in \Cref{fig:pipeline} (left).
We will then learn to classify $x_i^+$ and $x_i^-$ as true or false; if $x_i^+$ is true, then the answer to $q_i$ should be ``Yes'', and if $x_i^-$ is true, then the answer to $q_i$ should be ``No''.

In practice, we convert each task into a question-answering task with two possible labels, then we use task-specific zero-shot prompts to format questions and answers as strings to construct each contrast pair.
The opposite labels we use to construct contrast pairs can be ``Yes'' and ``No'' for a generic task, or they can be other tasks-specific labels, such as ``Positive'' and ``Negative'' in the case of sentiment classification. 
We describe the exact prompts we use to for each task in \Cref{append:prefix}.

\textbf{Feature extraction and normalization.}
Given a contrast pair $(x_i^+, x_i^-)$, CCS first computes the representations $\phi(x_i^+)$ and $\phi(x_i^-)$ using the feature extractor $\phi(\cdot)$.
Intuitively, there are two salient differences between $\phi(x_i^+)$ and $\phi(x_i^-)$: (1) $x_i^+$ ends with ``Yes'' while $x_i^-$ ends with ``No'', and (2) one of $x_i^+$ or $x_i^-$ is true while the other is false.
We want to find (2) rather than (1), so we first try to remove the effect of (1) by normalizing $\{\phi(x_i^+)\}$ and $\{\phi(x_i^-)\}$ independently.
In particular, we construct normalized representations $\tilde{\phi}(x)$ as follows:
\[
    \tilde{\phi}(x_i^+) := \phi(x_i^+) - \mu^+\,;\quad \tilde{\phi}(x_i^-) := \phi(x_i^-) - \mu^-,
\]
where $\mu^+, \mu^- \in \mathbb{R}^d$ are the means of $\{\phi(x_i^+)\}_{i=1}^n$ and $\{\phi(x_i^-)\}_{i=1}^n$. 
This normalization ensures that $\{\tilde{\phi}(x_i^+)\}$ and $\{\tilde{\phi}(x_i^-)\}$ no longer form two separate clusters.
In practice we also normalize the scale of the features, but this isn't essential for the method to work; see \Cref{appendix:normalization} for details.

\textbf{Mapping activations to probabilities.}
Next, we learn a probe $p_{\theta, b}(\tilde{\phi})$ that maps a (normalized) hidden state $\tilde{\phi}(x)$ to a number between $0$ and $1$ representing the probability that the statement $x$ is true. We use a linear projection followed by a sigmoid $\sigma(\cdot)$, i.e. $p_{\theta, b}(\tilde{\phi}) = \sigma(\theta^T \tilde{\phi} + b)$, but nonlinear projections can also work.
For simplicity, we sometimes omit the $\theta,b$ subscript in $p$. 

\textbf{Training objective.}
To find features that represent the truth, we leverage the consistency structure of truth. First, we use the fact that a statement and its negation should have probabilities that add up to $1$. This motivates the consistency loss:
\[
	L_{\text{consistency}}(\theta, b; q_i) := \left[p_{\theta, b}(x_i^+) - (1-p_{\theta, b}(x_i^-))\right]^2
\]

However, this objective alone has a degenerate solution: $p(x^+) = p(x^-) = 0.5$. To avoid this problem, we encourage the model to also be confident with the following confidence loss:
\[
	L_{\text{confidence}}(\theta, b; q_i) := \min\{p_{\theta, b}(x_i^+), p_{\theta, b}(x_i^-)\}^2
\]
We can equivalently interpret $L_{\text{confidence}}$ as imposing a second consistency property on the probabilities: the law of excluded middle (every statement must be either true or false).
The final unsupervised loss is the sum of these two losses, averaged across all contrast pairs:
\[
    L_{\text{CCS}}(\theta, b) := \frac{1}{n} \sum_{i=1}^n L_{\text{consistency}}(\theta, b; q_i) + L_{\text{confidence}}(\theta, b; q_i)
\] 
Note that both losses are necessary; $L_{\text{confidence}}$ alone also has a degenerate solution.

\textbf{Inference.} 
Both $p(x_i^+)$ and $1-p(x_i^-)$ should represent the probability that the answer to $q_i$ is ``Yes''. However, because we use a soft consistency constraint, these may not be exactly equal. To make a prediction on an example $x_i$ after training, we consequently take the average of these:
\[
    \tilde{p}(q_i) := \frac{1}{2}(p(x_i^+) + (1-p(x_i^-))
\]
We then predict that the answer to $q_i$ is ``Yes'' based on whether $\tilde{p}(q_i)$ is greater than $0.5$. 
Technically, we also need to determine whether $\tilde{p}(q_i) > 0.5$ corresponds to ``Yes'' or ``No,'' as this isn't specified by $L_{\text{CCS}}$.
For simplicity in our evaluations we take the maximum accuracy over the two possible ways of labeling the predictions of a given test set.
However, in \Cref{sec:appx:identifying} we describe how one can identify the two clusters without any supervision in principle by leveraging conjunctions.

%% file: sections/3-experiments.tex
\vspace{-5pt}
\section{Results}
\label{sec:experiments}
\vspace{-3pt}

\vspace{-2pt}
\subsection{Experimental Setup}\label{subsec:exp_setup}
\vspace{-5pt}
Here we give an overview of our experimental setup; see \Cref{sec:appx:implementation} for full details.
We provide code at \url{https://www.github.com/collin-burns/discovering_latent_knowledge}.

\textbf{Models.} We test six models: encoder-decoder models (T5 \citep{Raffel2020ExploringTL}, UnifiedQA \citep{Khashabi2020UnifiedQACF}, T0 \citep{Sanh2021MultitaskPT}), autoregressive models (GPT-J \citep{gpt-j}), and encoder-only models (RoBERTa \citep{Liu2019RoBERTaAR}, DeBERTa \citep{He2021DeBERTaDB}).

\textbf{Data.} We test models on $10$ datasets: sentiment classification (IMDB \citep{maas2011learning} and Amazon \citep{mcauley2013hidden}), topic classification (AG-News \citep{zhang2015character} and DBpedia-14 \citep{lehmann2015dbpedia}), NLI (RTE \citep{wang2018glue} and QNLI \citep{rajpurkar2016squad}), story completion (COPA \citep{roemmele2011choice} and Story-Cloze \citep{mostafazadeh2017lsdsem}), question answering (BoolQ \citep{clark2019boolq}), and common sense reasoning (PIQA \citep{bisk2020piqa}).

We convert each dataset to a yes-no question-answering task or a binary classification task, as described in \Cref{sec:appx:implementation}. 
We balance the labels and randomly subsample 1000 examples from each dataset (except for COPA, which has only 500 examples total), then randomly split each dataset into an unsupervised training set ($60\%$ of the data) and test set ($40\%$). 
We subsample each dataset for computational efficiency reasons; because we aggregate over ~9 prompts per dataset, 10 datasets, and 6 models, 1000 datapoints per dataset actually corresponds to approximately 180k examples in total.

\textbf{Methods.} We test four main methods: zero-shot, calibrated zero-shot, Contrast-Consistent Search (CCS), and Logistic Regression (LR). 
Zero-shot works by predicting the answer with the highest log probability according to the language model, averaged across the tokens that make up that label.
Calibrated zero-shot works by balancing zero-shot predictions to be $50/50$ for each answer, as we describe in more detail below, similar to \citet{Zhao2021CalibrateBU}.
For Logistic Regression we train on the training split for each dataset using $(\tilde{\phi}(x^+) , \tilde{\phi}(x^-))$ as the covariates, then evaluate on the corresponding test split.
We treat LR as a ceiling since it uses labeled data. 

When testing CCS, we optimize it 10 times using AdamW \citep{Loshchilov2017FixingWD} with learning rate $0.01$, then take the run with the lowest unsupervised loss. Unless otherwise specified, we train CCS using all prompts for a single training set (normalized independently for each prompt), then evaluate it on the corresponding test split.

\textbf{Zero-shot baselines.} 
Zero-shot outputs sometimes suffer from miscalibration \citep{Zhao2021CalibrateBU}, in which models are biased towards predicting specific answers.
Calibrating the outputs to be uniform over different answers can mitigate this problem. 
We use a variant of the calibration method presented in \citet{Zhao2021CalibrateBU} by balancing predictions to be 50/50 across the two output labels. Specifically, if $l_+$ and $l_-$ are the logits for the positive and negative label respectively, then instead of classifying an example as positive if $l_+ > l_-$, we classify it as positive if $l_+ > l_- + \gamma$, where we select the threshold $\gamma \in \mathbb{R}$ so that the predictions are balanced. 
We find this increases accuracy by about $5\%$ on average. 
Unless otherwise specified, we always report zero-shot accuracy after calibration.

Encoder-only models (e.g. RoBERTa and DeBERTa) cannot be easily used to do zero-shot classification out of the box, so to evaluate them we follow the method of \citet{Yin2020UniversalNL}: we finetune both models on an NLI dataset (MNLI, which we do not evaluate on) and treat the difference between the entailment and contradiction probabilities as the effective logit. This provides a strong zero-shot baseline for encoder-only models that works even for non-NLI tasks \citep{Yin2020UniversalNL}. This finetuning isn't necessary for CCS to work on encoder-only models (see \Cref{appx:sec:mlm_results}), but we test CCS using the same MNLI-finetuned models for ease of comparison.

\textbf{Hidden states.} 
We extract the hidden states corresponding to the last token in the last layer of each model for simplicity, unless otherwise specified. 
For encoder-decoder models, we evaluate CCS on the last layer hidden states of both the encoder and decoder, and use whichever one generally achieves a lower unsupervised loss; for T0 this is the decoder hidden states, while for T5 and UnifiedQA this is the encoder hidden states.
See \Cref{appendix:contrast_pairs} for further implementation details, such as tokenization.

\textbf{Prompts.} To reduce prompt sensitivity, we use between $8$ and $13$ prompts for each dataset ($9$ on average), derived or slightly modified from \citet{Sanh2021MultitaskPT}.
Unless otherwise specified, we average across all prompts when showing results.
To construct contrast pairs, we let $x_i^+$ be the zero-shot prompt using $q_i$ and the first label (e.g.~``Positive'' for sentiment classification datasets) and let $x_i^-$ be the prompt using $q_i$ and the second label (e.g.~``Negative'').
We describe all prompts in \Cref{sec:appx:setup}.

\begin{table*}[t]
	\centering
    \small
    \begin{tabular}{l|cccccc|c}
	Method  & RoBERTa & DeBERTa & GPT-J & T5 & UQA & T0$^*$ &  Mean$^*$\\
    \hline
    0-shot & 60.1(5.7) & 68.6(8.2) & 53.2(5.2) & 55.4(5.7) & 76.8(9.6) & 87.9(4.8)  &  62.8(6.9) \\
    Calibrated 0-shot & \textbf{64.3(6.2)} & 76.3(6.0) & 56.0(5.2) & 58.8(6.1) & 80.4(7.1) & 90.5(2.7) &   67.2(6.1)\\
    CCS  & 62.1(4.1) & \textbf{78.5(3.8)} & 61.7(2.5) & 71.5(3.0) & 82.1(2.7) & 77.6(3.3) &  71.2(3.2)\\
    CCS (All Data)  & 60.1(3.7) & 77.1(4.1) & \textbf{62.1(2.3)} &\textbf{72.7(6.0)} & \textbf{84.8(2.6)} & 84.8(3.7) &  \textbf{71.5(3.7)}\\
    \hline
    LR (Ceiling) & 79.8(2.5) & 86.1(2.2) & 78.1(2.3) & 84.6(3.1) & 89.8(1.9) & 90.5(2.1) & 83.7(2.4)\\
    \end{tabular}
	\caption{
	Accuracy of each method and model averaged across all prompts and dataset, with the average standard deviation of accuracy across different prompts shown in parentheses. 
	For most models, CCS outperforms zero-shot accuracy and exhibits lower sensitivity to prompts, even though this was not our goal.
    This shows that we can recover knowledge from language model activations without supervision, and can do so in a way that is competitive with strong baseline methods that use model outputs.
    $^*$T0 was trained on 9 out of 10 of the datasets we evaluate on, including some of the data in our test splits, so we ignore it when averaging over models.
	}
    \label{tab:avg}
    \vspace{-12pt}
\end{table*}

\vspace{-8pt}
\subsection{Evaluating CCS}
\label{sec:exp_mainsub}
\vspace{-5pt}

\subsubsection{CCS Outperforms Zero-Shot}
\label{subsec:ccs_outperforms}
\vspace{-5pt}
We evaluate CCS on all 6 models and compute the average accuracy across all datasets and prompts. 
T0 was trained on 9 out of 10 of the datasets we evaluate on, including some of the data in our test splits, so we ignore it when averaging over models to avoid unfair comparisons.
We display the results in \Cref{tab:avg}.
To assess prompt sensitivity, for each model and dataset we compute the standard deviation (s.d.) of accuracy across different prompts, then average the resulting standard deviations across all datasets, which we show in parentheses in \Cref{tab:avg}.
For comparison, we also include results when training CCS on all datasets simultaneously, which we refer to as CCS (All Data).

CCS attains an accuracy of $71.2\%$ on average, compared to $67.2\%$ for calibrated zero-shot. 
It outperforms zero-shot accuracy for every model, except for RoBERTa (where it does $2\%$ worse) and T0 (for which zero-shot accuracy is inflated).
Training on all datasets improves accuracy by only an insignificant amount on average ($0.3\%$), but with large gains for T0 in particular ($77.6\% \rightarrow 84.8\%$).

These results show that CCS can \textit{exceed} the performance of strong baseline methods that access the model outputs, even though this wasn't our main goal.
This indicates that it is indeed possible to classify examples with high accuracy using only unlabeled model representations.

\vspace{-5pt}
\subsubsection{CCS Is Robust To Misleading Prompts}
\label{subsec:ccs_robust}
\vspace{-3pt}
Recall our goal: to discover latent knowledge in a language model even when the model outputs false text. 
In particular, language models are typically trained to imitate text whether or not it is correct, so if a model sees false text it should intuitively be more likely to predict that subsequent text will also be false.
Based on this idea, we provide an initial proof of concept that CCS can make progress toward our goal by constructing prompts that aim to mislead the outputs of language models.
Specifically, we add a prefix to the beginning of our zero-shot prompts that consists of questions answered incorrectly (\Cref{fig:prefix}). 
The hope is that such a prefix will decrease zero-shot accuracy because the model will imitate its context and answer subsequent questions incorrectly even if it internally ``knows'' better.\footnote{In practice we found that model behavior was qualitatively similar between using this prefix with correct and incorrect answers, in agreement with the findings of \citet{Min2022RethinkingTR} that using incorrect labels in demonstrations often does not significantly degrade the performance of a few-shot prompt. Consequently, the prefix may instead actually be reducing accuracy because it is out-of-distribution. See also \citet{Kim2022GroundTruthLM} for more experiments on the subtleties of correct vs incorrect labels for few-shot prompts.}
We found that while most models are robust to this type of prefix (see \Cref{append:prefix}), it significantly drops calibrated zero-shot performance in UnifiedQA, decreasing accuracy from $80.4\%$ to $70.9\%$.

We evaluate CCS on these examples and show the results in \Cref{fig:acc_drop} of the Appendix.
We find that despite the $9.5\%$ drop in zero-shot accuracy, CCS maintains high accuracy ($82.1\% \rightarrow 83.8\%$). 
This provides evidence that our method can still work well even when model outputs are unreliable.

\vspace{-5pt}
\subsection{Analyzing CCS}\label{sec:understanding_cfs}
\vspace{-3pt}
We have shown that CCS attains strong classification performance in standard settings and when we deliberately mislead models.
Moreover, we have described our motivation as discovering latent representations of truth in language models, but in practice CCS just finds a direction in representation space that attains high accuracy.
This raises the question: in what sense is CCS actually finding ``truth'' features?
We now provide a preliminary investigation of this question.

\vspace{-5pt}
\subsubsection{CCS Finds A Task-Agnostic Representation of Truth} 
\label{subsec:transfer}
\begin{figure*}[t]
    \centering
    \includegraphics[width=\textwidth]{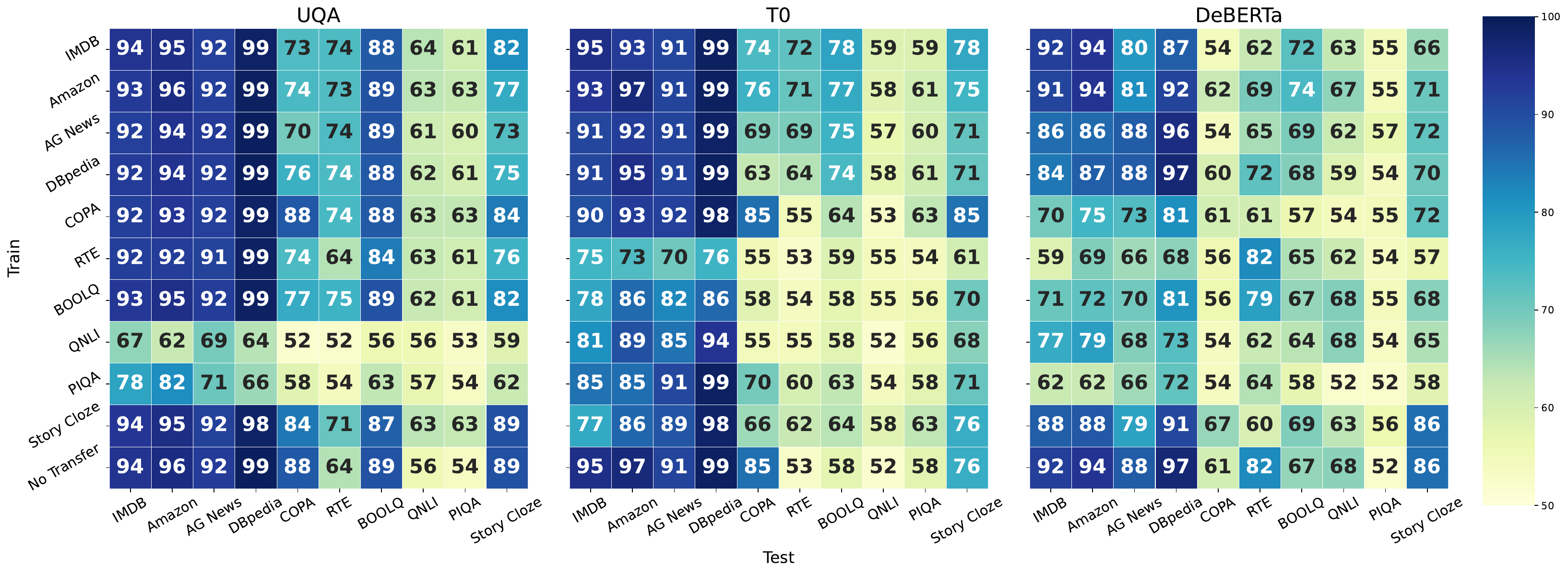}
    \caption{Transfer Performance using CCS on UnifiedQA, T0 and DeBERTa. 
	The y-axis corresponds to the training dataset, and the x-axis corresponds to the test dataset.
	The final row (``No Transfer'') corresponds to training and testing on the same dataset, which is the same as the diagonal.
	On most datasets, CCS transfers well to other datasets (relative to no transfer), including to different tasks with completely different labels.
    In some cases transfer even outperforms the no-transfer setting.
    See \Cref{sec:appx:transfer} for results with other models.
    }
    \label{fig:transfer}
    \vspace{-10pt}
\end{figure*}

\vspace{-3pt}
From the results described so far, it may be possible that the classifier we find is capturing dataset-specific properties, such as which of two labels (e.g. ``Yes'' vs.~``No'') is more likely to be correct. 
We rule this out by showing that it generalizes across completely different tasks, including ones with different label spaces (such as from ``Yes'' and ``No'' for a generic task to ``Positive'' and ``Negative'' for sentiment classification). 
In particular, we train and test CCS on every pair of datasets, and show the resulting transfer for several models in \Cref{fig:transfer}.
(See \Cref{append:sec:complete_intermediate} for results with other models.)

We find that CCS indeed transfer wells: in the majority of datasets, the transfer accuracy of CCS is competitive with training and testing on the same dataset (the last row in \Cref{fig:transfer}, ``No transfer''). 
Transfer performance can even outperform the no-transfer setting, especially when we train on simpler tasks, such as sentiment classification.
For example, training CCS on the Amazon sentiment dataset achieves an average transfer accuracy of $71.8\%$, which is $0.6\%$ higher than CCS without transfer.
We speculate that this performs so well because the difference in representations between correct and incorrect answers is especially pronounced for easier tasks like Amazon. 

Additionally, transfer accuracy tends to be similar for many datasets.
For example, training on IMDB (row 1 of \Cref{fig:transfer}) has similar accuracy as training on DBPedia (row 4).
This provides evidence that CCS can find a functionally similar direction across many different types of training datasets. 
Overall, these results suggest that (1) models may have a task-agnostic representation related to what is true, and that (2) CCS may approximately find this representation even without diverse data.

\vspace{-5pt}
\subsubsection{CCS Does Not Just Recover Model Outputs} 
\label{subsec:not_model_outputs}
\vspace{-3pt}
One possibility is that CCS can only recover knowledge already contained in a model's outputs.
We have shown that CCS can outperform zero-shot accuracy (\Cref{subsec:ccs_outperforms}), especially when model outputs are misled (\Cref{subsec:ccs_robust}), which already provides evidence against this possibility.
We now provide additional evidence against this concern.

First, if CCS were just recovering knowledge in the model outputs, using the last layer of a network (right before the outputs) should presumably outperform intermediate layers (which are more causally distant from the outputs). However, for T5 and UnifiedQA, we find that using hidden states in the middle of the network outperform hidden states at the end of the network when using CCS (see \Cref{fig:crosslayer-CCS-6model}). This is especially true for UnifiedQA on misleading prefixes; we find that using the encoder hidden states is robust to misleading prefixes (\Cref{subsec:ccs_robust}), but that accuracy using the decoder hidden states drops from $81.0\%$ to $73.5\%$, a similar amount to zero-shot accuracy.
This suggests that compared to the later layers of a model, intermediate layers are more robust and less correlated with the model outputs, and that CCS can take advantage of this.

Finally, if CCS were just recovering knowledge in the model outputs, we would only expect it to work in cases where model outputs are informative. However, we show in \Cref{appx:sec:mlm_results} that CCS still works with masked language models when their outputs are uninformative: when we don’t [MASK] any input tokens, and when we prompt models so that the labels used to construct contrast pairs appear in the \textit{middle} of a prompt rather than at the end.
These results show that CCS can sometimes recover latent knowledge in a model that is distinct from---and more useful than---what the model outputs.

\vspace{-5pt}
\subsubsection{Truth is a Salient Feature}
\label{sec:CRC}
\vspace{-3pt}

From the results we have presented so far, it is possible that the direction learned by CCS is difficult to find and requires using a large amount of unsupervised data.
We provide evidence against this possibility by showing that finding such a direction can both (1) often be done with only a small amount of data, and can also (2) often be done by essentially taking the top principal component of a slightly modified representation space.

\begin{figure}[t]
    \centering
     \includegraphics[width=0.54\textwidth]{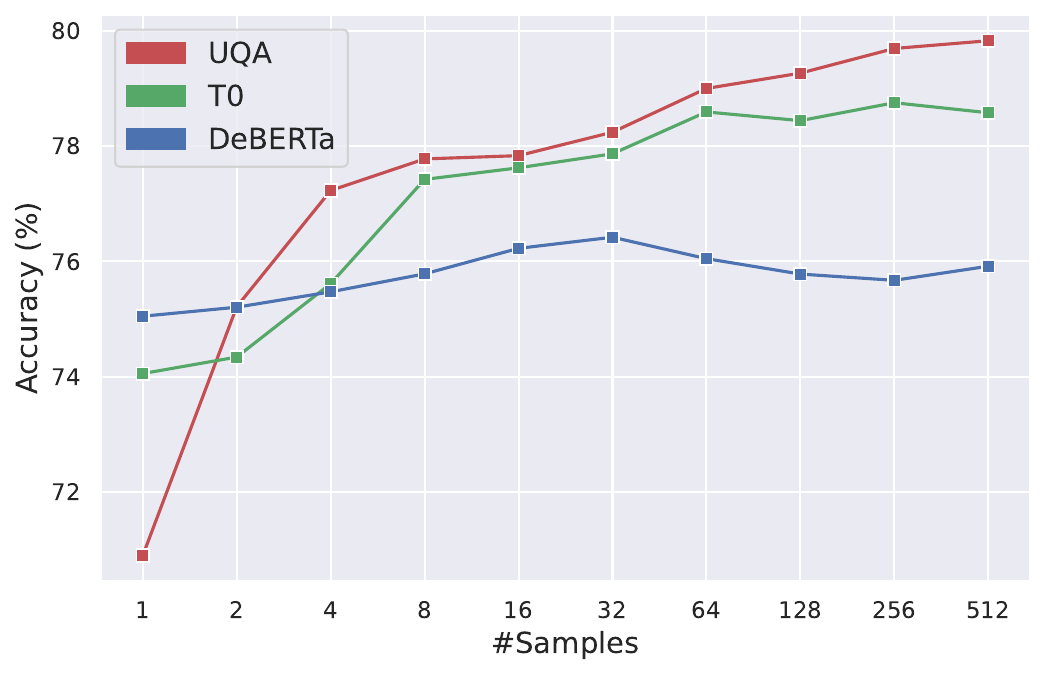}
    \caption{Accuracy when we train CCS on $k$ samples for different values of $k$ (each time averaged across 32 trials). We use the single prompt with the highest zero-shot accuracy for each dataset and model. While CCS benefits from more examples, it can often work well with limited data.}
    \label{fig:numdata}
     \vspace{-12pt}
\end{figure}

\textbf{CCS doesn't require much data.}
We now evaluate how well CCS performs with different amounts of data. 
Whereas before we trained CCS using the full training set and all prompts, here we use limited data and a single prompt. 
Specifically, we train using only $k$ unlabeled contrast pairs, using the single prompt for each model and dataset that achieves the highest zero-shot accuracy. %
We still test on all prompts for each dataset.
We resample $k$ points $32$ times for each of $k=1,2,4,\cdots$, and take the average accuracy across those $32$ samples.
Finally, we plot the average such accuracy across all datasets and prompts for several models in \Cref{fig:numdata}.

We find that while CCS benefits from more data, it can often do well with very limited data.
In fact, it can sometimes even do well with only a single contrast pair, though we find high variance across individual datasets; see \Cref{append:complete_sample_results} for more details.
This suggests that the strong performance of CCS does not primarily come from using a large amount of unsupervised data, and indicates that the direction learned by CCS may be relatively easy to find.

\textbf{Contrastive Representation Clustering.}
We now show that directions correlated with the truth may be ``salient'' in a different way: by showing that we can also find such directions using either (1) PCA or (2) clustering.
Specifically, suppose we construct contrast pairs ($x^+_i$, $x^-_i$) as before. 
Intuitively, these two examples are qualitatively almost identical except that one is true and the other is false, so the main difference between the representations $\tilde{\phi}(x_i^+)$ and $\tilde{\phi}(x_i^-)$ should relate to truth.
Consequently, we can take the differences in (normalized) hidden states, $\{\tilde{\phi}(x_i^+) - \tilde{\phi}(x_i^-)\}_{i=1}^n$, and cluster them.
We call this method Contrastive Representation Clustering (CRC). 
Clustering can be achieved by taking the top principal component (TPC) and thresholding at $0$, or by doing a ``bimodal salience search'' (BSS) to find a direction that looks bimodal; see \Cref{appendix:CRC} for further details.

We compare CCS and these two variants of Contrastive Representation Clustering in \Cref{tab:CFS+CRC}, using the same setting as in \Cref{tab:avg}.
While CCS performs best, all methods attain high accuracy and are competitive with zero-shot performance.
This indicates both that (1) representations of truth often lie in a high-variance direction in the contrastive representation space $\{\tilde{\phi}(x_i^+) - \tilde{\phi}(x_i^-)\}_{i=1}^n$, and also that (2) true and false examples are often well-clustered in this same contrastive space.
This strengthens the idea that representations of truth may be salient features inside models that are relatively easy to find. This may help explain why CCS can perform well without using any supervision, and how it can do so even with only a limited amount of unlabeled data.

\begin{table}[b]
	\centering
    \small
    \vspace{-8pt}
    \begin{tabular}{l|cccccc|c}
	Method & RoBERTa & DeBERTa & GPT-J & T5 & UQA & T0$^*$  & Mean$^*$\\
    \hline
    Calibrated 0-shot & 64.3(6.2) & 76.3(6.0) & 56.0(5.2) & 58.8(6.1) & 80.4(7.1) & 90.5(2.7) &   67.2(6.1)\\
    CCS & 62.1(4.1) & \textbf{78.5(3.8)} & 61.7(2.5) & \textbf{71.5(3.0)} & \textbf{82.1(2.7)} & 77.6(3.3) & \textbf{71.2(3.2)}\\
    CRC (TPC) & \textbf{65.7(4.9)} & 77.0(4.9) & 60.8(3.2) & 68.3(4.6) & 78.8(3.0) & 65.3(6.8) & 70.1(4.1) \\
    CRC (BSS) & 63.6(5.5) & 77.9(4.9) & \textbf{61.9(2.3)} & 67.4(2.2) & 80.0(3.4) & 79.0(5.1) & 70.2(3.7) \\

    \end{tabular}
	\caption{
    We compare CCS to two variants of Contrastive Representation Clustering: TPC, which clusters by projecting onto the top principal component, and BSS, which clusters by finding a direction that looks bimodal.
	We show accuracy and standard deviation of each model averaged across all prompts and datasets, in the same setting as \Cref{tab:avg}.
    We find that CCS generally performs the best, but that all methods are competitive with zero-shot.
	}
    \label{tab:CFS+CRC}
    \vspace{-8pt}
\end{table}

%% file: sections/4-related.tex
\vspace{-8pt}
\section{Related Work}
\label{sec:related_work}
\vspace{-5pt}

\textbf{Zero-Shot Prompting.}
Since the release of GPT-3 \citep{Brown2020LanguageMA}, one of the main paradigms for eliciting what models know has been zero-shot prompting \citep{Liu2022PretrainPA,Beltagy2022ZeroAF}. 
Zero-shot exploits how language models are trained to predict diverse data from the internet, which incidentally includes tasks such as question-answering.
If prompted appropriately, this can be used to solve various useful tasks with reasonable performance \citep{Brown2020LanguageMA}.
However, these models are trained to imitate human-generated data, which bounds the quality of their outputs.

Many methods improve upon vanilla zero-shot prompting \citep{Liu2022PretrainPA,Zhao2021CalibrateBU, Lu2022FantasticallyOP,Wei2022ChainOT,Min2022MetaICLLT}.
While our goal is not to improve zero-shot performance, some of the ideas underlying these methods are similar to CCS.
Particularly relevant are methods that also leverage unsupervised consistency properties, such as \citet{Jung2022MaieuticPL,Zhou2022PromptCF}.
However, these methods still bootstrap from language model outputs trained via imitation learning, which limits their applicability to our main goals.

To illustrate this, imagine we train reinforcement learning agents to play a game such as Diplomacy \citep{Bakhtin2022HumanlevelPI}, in which players have incentives to lie to each other.
Then those agents may learn to lie in a way that is difficult to detect, but they may still internally represent whether they are lying.
Because their outputs would be deliberately misleading, standard zero-shot methods may be very unreliable.
In contrast, techniques like CCS may still be able to detect whether those models are lying by finding representations of truth in their activations that contradict their outputs.

\textbf{Truthfulness.}
There has been increasing interest in making language models truthful \citep{Evans2021TruthfulAD,Lin2022TruthfulQAMH}.
One aspect of truthfulness that has received substantial attention is factuality \citep{Thorne2018FEVERAL,Maynez2020OnFA}. For instance, many techniques aim to improve the factuality of models by augmenting them with retrieval methods \citep{Nakano2021WebGPTBQ,Menick2022TeachingLM}, which allows them to cite their sources.
In contrast, we focus on truthfulness more generally, which also includes procedural knowledge such as reasoning or natural language inference tasks.

An approach to making language models truthful in this more general setting is to finetune them using either human demonstrations \citep{Khashabi2020UnifiedQACF,Sanh2021MultitaskPT,Zhong2021AdaptingLM,Wei2022FinetunedLM} or reinforcement learning from human feedback \citep{Christiano2017DeepRL,Stiennon2020LearningTS,Askell2021AGL,Bai2022TrainingAH,Ouyang2022TrainingLM}.
These techniques have been widely successful at improving performance, but unlike our method they rely on being able to provide ground truth labels, which can be intractable in many settings.

Some work has aimed to go beyond the direct supervision humans can provide by augmenting supervision with AI systems \citep{Christiano2018SupervisingSL,Irving2018AISV,Leike2018ScalableAA,Perez2022RedTL}.
This may expand the range of applications we can supervise, but many of these proposals remain theoretical, and it is unclear just how far these techniques can generalize.
\citet{ELK} reframes this issue by posing the problem of Eliciting Latent Knowledge (ELK). 
Like our problem statement, ELK is about eliciting knowledge from models even in cases where humans cannot evaluate that knowledge.
However, ELK frames this as a worst-case theoretical problem, while we frame this as an empirical problem that we can make progress on using current models.

%% file: sections/5-conclusion.tex
\vspace{-8pt}
\section{Discussion}
\vspace{-5pt}
\subsection{Limitations and Future Work} 
\vspace{-3pt}

Our work has a number of limitations.
First, CCS relies on the existence of a direction in activation space that separates true and false inputs well, in the sense that a supervised probe on the activations would be able to attain high accuracy (if it hypothetically had access to the ground truth labels).
This requires that a model is both \textit{capable} of evaluating the truth of a given input, and also that the model \textit{actively evaluates the truth} of that input. 
It is not clear when these conditions hold precisely.

Second, we did not evaluate our method on setups involving active ``lying'' or ``deception'' \citep{Kenton2021AlignmentOL,Evans2021TruthfulAD} by models, as we aren't aware of existing evaluation setups for this setting. 
If future work develops such a setup, a good stress test would be to apply CCS to do ``lie detection'' in that setting. 
This may require modifications or extensions to the method, such as more explicitly ensuring that it recovers the truth of an input rather than what the model says.

There are also various straightforward improvements to our method that one could explore.
This includes adding additional consistency constraints, improving its reliability, calibrating its probabilities, generalizing it beyond the yes-no question-answering setting, generalizing it to cases where answers aren't clear-cut, and closing the remaining gap between CCS and the logistic regression ceiling.

\vspace{-5pt}
\subsection{Conclusion} 
\vspace{-3pt}
As language models become more capable, they will be increasingly used as components in larger AI systems trained with reinforcement learning. 
As this occurs, falsehoods arising from misaligned training objectives may become more common, severe, and difficult to detect.
In principle, models may even develop instrumental incentives to lie: for example, if a human evaluator would disapprove of bad model behavior, models may learn to lie about their behavior to achieve higher reward.
If so, those models would be optimizing against human evaluators, making it precarious to rely on those evaluators for assessing the truth of what models say.
Because unsupervised methods for eliciting answers circumvent this issue, they may still work even in this scenario.
We show that it is possible to make progress on such methods today; the empirical success of our approach suggests that unsupervised methods are both a tractable and underexplored research direction.

%% file: sections/6-appendix.tex
\clearpage
\section{Identifying Clusters}
\label{sec:appx:identifying}
CCS predicts yes-no answers to a set of unlabeled questions $q_1, \ldots, q_n$. 
However, it does not identify which prediction label corresponds to true and which corresponds to false. 
We now describe how one can do so in principle, as long as a model is also able to take conjunctions.

Suppose we run our method and find, for example, that $x$ and $x'$ end up in different clusters. 
Then we know that these two statements have opposite truth values.
As a result, if we take the conjunction of these two statements, $x \wedge x'$ then it should be false, and if we take the disjunction, $x \vee x'$, then it should be true.
This allows us to identify which cluster is true and which is false in a completely unsupervised way.

\begin{figure}
    \vspace{-10pt}
    \centering
    \includegraphics[width=0.5\textwidth]{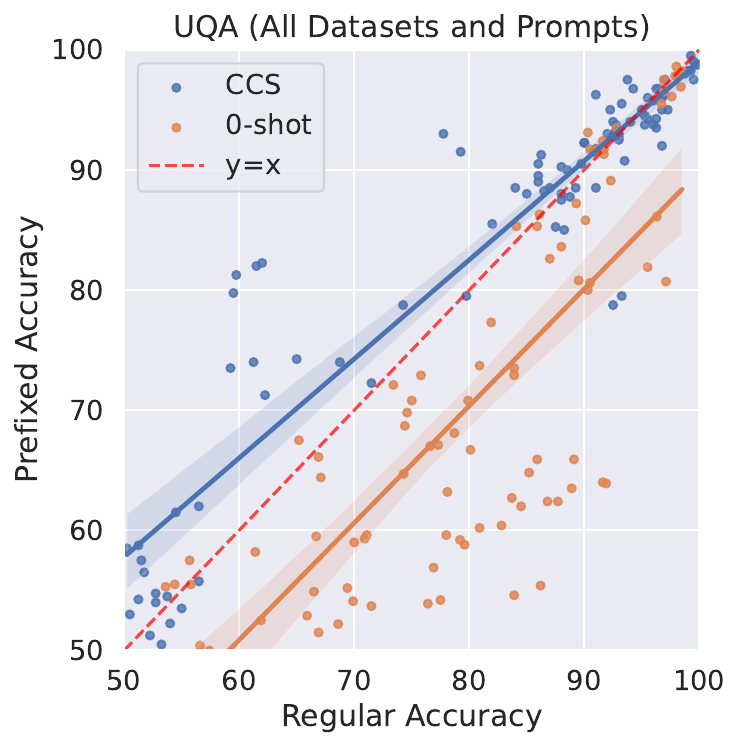}
    \caption{
    Zero-shot accuracy drop caused by the misleading prefix shown in \cref{fig:prefix} for UnifiedQA. Each point corresponds to one dataset and one prompt. 
    On average, zero-shot accuracy drops $9.5\%$, while CCS performs $1.7\%$ better.
    This suggests that the latent knowledge inside a model that CCS recovers can be robust even when model outputs become less reliable.
    }
    \label{fig:acc_drop}
    \vspace{-10pt}
\end{figure}

\section{Misleading Prefix Details}
\label{append:prefix}
\subsection{Misleading Prefix}
\begin{figure}[h]
	\begin{center}
	\includegraphics[width=0.45\textwidth]{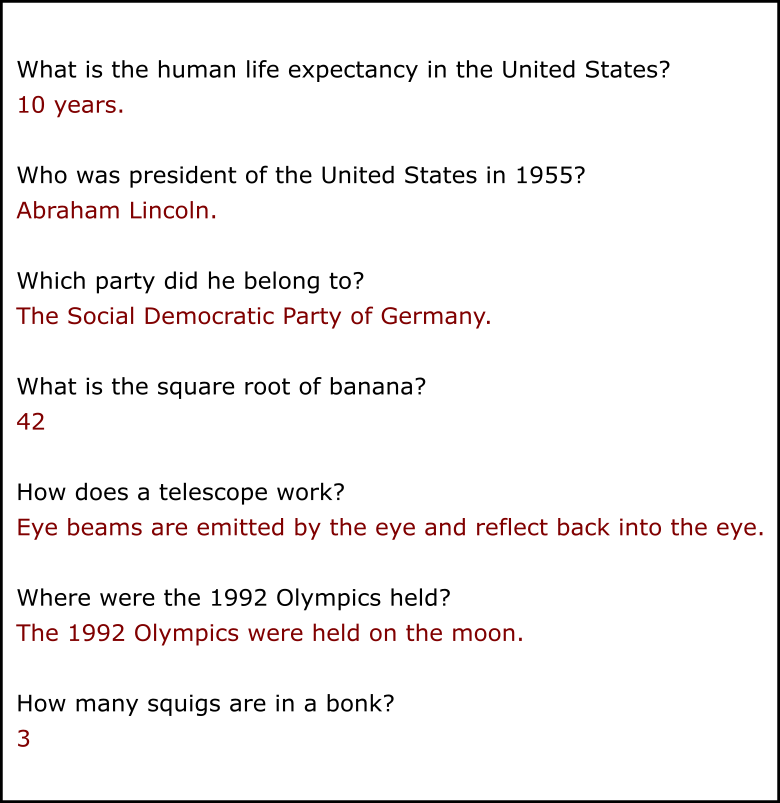}
	\end{center}
	\caption{
    The added prefix used to mislead models in a way that is egregiously incorrect or confusing. 
	}
	\label{fig:prefix}
\end{figure}

In \cref{sec:exp_mainsub}, we mentioned that model outputs can be biased by using the misleading few-shot prefix shown in \Cref{fig:prefix}. 
We show the effect of the misleading prefix for all models in \Cref{tab:prefix}, and we illustrate the drop in accuracy for each dataset and prompt in \Cref{fig:acc_drop}.

\subsubsection{Evaluating The Effect of The Misleading Prefix}
\begin{figure*}[h]
	\begin{center}
	\includegraphics[width=0.98\textwidth]{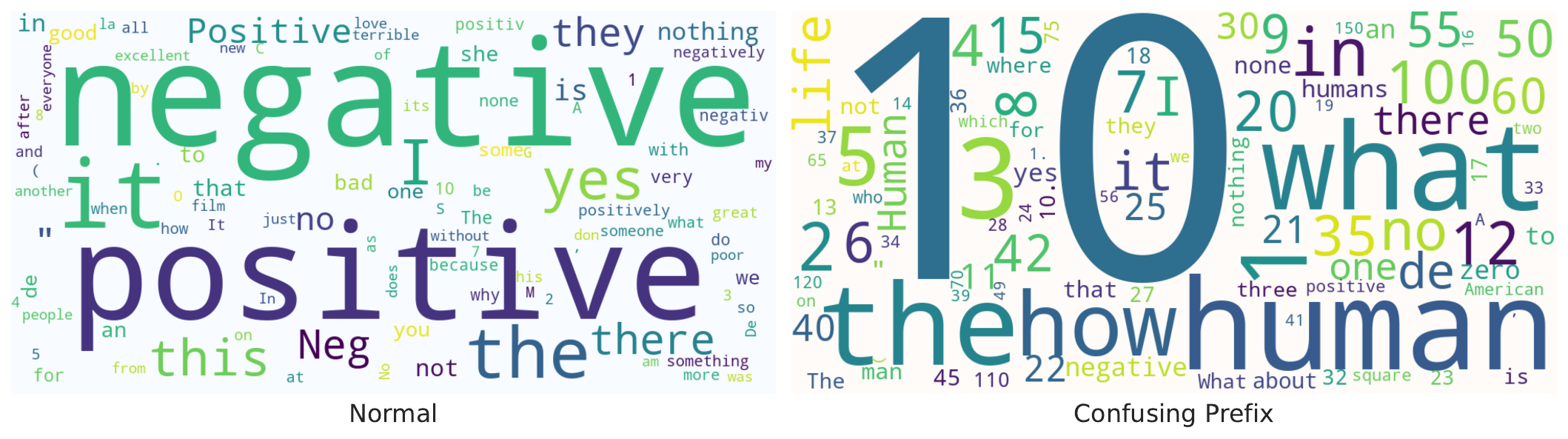}
	\end{center}
	\vspace{-4mm}
	\caption{
    The top $100$ tokens with highest probability averaged across examples in IMDB for UQA.
    Without the prefix (left), the actual labels (``positive'' and ``negative'') and their synonyms (e.g. ``Positive'' and ``Neg'') have high probability.
    With the prefix, the model's output becomes mostly irrelevant to sentiment; the actual labels are in the top $100$, but no other synonyms are.
	}
	\label{fig:wordcloud}
\end{figure*}

The hope is that the misleading prefix causes models to imitate the sorts of incorrect answers in the prefix. 
However, we found that results are qualitatively similar with correct answers, making the actual interpretation of the effect of this prefix ambiguous.
Nevertheless, to better understand what this prefix is doing, we visualize the top $100$ tokens for both this misleading prefix and no prefix on average across all examples for UnifiedQA on the IMDB sentiment dataset, and show the results in \Cref{fig:wordcloud}. 
Without this prefix, the top two tokens are the actual labels (`positive’ and `negative’). In contrast, with the prefix, these tokens have lower probabilities relative to other tokens, with the highest probability tokens instead being irrelevant to the true answer. 
For example, since the prefix includes several answers with numbers, many of the top tokens are also numbers. 
This provides some evidence that the prefix is actually causing the model to output false text by imitating its context in a meaningful sense.

\begin{table*}[b]
	\centering
    \small
    \begin{tabular}{ll|cccccc|c}
 
	Method & Prefix & RoBERTa & DeBERTa & GPT-J & T5 & UQA & T0$^*$ &  Mean$^*$\\
    \hline
    \multirow{2}*{\makecell[l]{Calibrated\\0-shot}} & Regular & 64.3(6.2) & 76.3(6.0) & 56.0(5.2) & 58.8(6.1) & 80.4(7.1) & 90.5(2.7) & 67.2(6.1) \\
 & Prefix & 65.6(5.3) & 75.5(6.1) & 59.2(4.6) & 56.0(4.0) & 70.9(7.8) & 88.2(4.2) & 65.4(5.6) \\
 \hline
 \multirow{2}*{CCS} & Regular & 62.1(4.1) & 78.5(3.8) & 61.7(2.5) & 71.5(3.0) & 82.1(2.7) & 77.6(3.3) & 71.2(3.2) \\
 & Prefix & 62.2(3.6) & 75.4(5.2) & 61.2(1.8) & 73.2(2.6) & 83.8(2.4) & 75.0(2.7) & 71.2(3.1) \\
 \hline
\multirow{2}*{TPC} & normal    & 65.7(4.9) & 77.0(4.9) & 60.8(3.2) & 68.3(4.6) & 78.8(3.0) & 65.3(6.8) & 70.1(4.1) \\
 & confusion & 65.6(5.0) & 77.4(5.4) & 61.1(2.1) & 69.4(4.2) & 76.7(3.4) & 64.6(7.7) & 70.0(4.0) \\
\hline
\multirow{2}*{BSS} & normal    & 63.6(5.5) & 77.9(4.9) & 61.9(2.3) & 67.4(2.2) & 80.0(3.4) & 79.0(5.1) & 70.2(3.7) \\
 & confusion & 62.2(4.7) & 76.5(5.8) & 62.0(2.7) & 67.2(2.2) & 83.4(3.4) & 74.6(3.5) & 70.3(3.7) \\
 \hline
 \multirow{2}*{LR} & Regular & 79.8(2.5) & 86.1(2.2) & 78.1(2.3) & 84.6(3.1) & 89.8(1.9) & 90.5(2.1) & 83.7(2.4) \\
 & Prefix & 79.5(2.7) & 86.3(2.6) & 79.1(2.9) & 86.3(2.8) & 90.1(2.2) & 90.3(2.3) & 84.3(2.6) \\

    \end{tabular}
	\caption{
	Accuracy of each method and model averaged across all prompts and datasets, with the same setting as \cref{tab:avg}.
    ``Regular'' means no prefix is added, while ``Prefix'' corresponds to text in \cref{fig:prefix}. Notice that while the 0-shot accuracy decreases due to the prefix, all of our methods are more resistant to the influence. 
	}
    \label{tab:prefix}
    \vspace{-10pt}
\end{table*}

\clearpage
\section{Masked Language Modeling Results}
\label{appx:sec:mlm_results}
We now provide an initial demonstration that CCS can work well even when a model's outputs are not very useful.

First, we evaluate our method on a model trained exclusively with the masked language modeling (MLM) objective: DeBERTa-v2 \citep{He2021DeBERTaDB} \textit{without} NLI finetuning (unlike in \Cref{sec:experiments}, where we finetuned DeBERTa on an NLI task to use it in the zero-shot setting). 
The outputs of DeBERTa are unlikely to be meaningful if we provide a raw input without using any [MASK] tokens\footnote{The MLM objective first selects 15\% of tokens to mask. Of the masked tokens, 80\% are replaced with [MASK], 10\% are randomly replaced with a different token, and 10\% are left unchanged \citep{He2021DeBERTaDB}. As a result, for non-[MASK] tokens, the most likely output token should usually be the original input token itself, which is uninformative.}. 
If CCS only works when the model outputs are useful, then it should perform poorly in this setting.

We also consider one other way that model outputs can be uninformative. Given a sequence of tokens as input, DeBERTa has a different set of output logits for each input token. For a given contrast example (a question and a candidate answer), the most informative output should intuitively be the logits that predict the answer. For instance, if the contrast example $x^+$ is ``Is 2+2=4? Yes'', then the most informative output should be the logits corresponding to candidate answer ``Yes''. Consequently, if we instead format each example so that the label (e.g. ``Yes'' or ``No'') appears in the \textit{middle} of the prompt, then the output corresponding to the \textit{final} token should not be very informative. If CCS only works well when the model outputs are useful, then it should also perform poorly in this setting (as long as CCS uses the last-token hidden states as usual).

To show that CCS can work even when model outputs aren't informative, based on the above discussion we now do an initial test of CCS when we simultaneously (1) use DeBERTa-v2 (MLM-pretrained only), and (2) format inputs so that the label is in the middle of the prompt rather than at the end. In particular, we apply CCS on the Amazon dataset, where we randomly sample 1000 new (unlabeled) points and use a 60/40 train-test split as before. As usual we continue to use the last-token hidden states for CCS. As usual, we use the default Huggingface tokenizer and we do not [MASK] any input tokens. We use the following custom prompt to test (2):

\begin{verbatim}
The following movie review expresses a [label] sentiment:\n[text]
\end{verbatim}

Even though it is unclear why the model outputs should be useful with this setup, we find that CCS can indeed still perform well, attaining an accuracy of $93.7\%$. For comparison, this is nearly identical to the approximately $94\%$ accuracy of CCS when evaluated on the Amazon dataset using NLI-finetuned DeBERTa and the original prompts (see \Cref{fig:transfer_all}).

We argued above that the model outputs should not be very useful in this setting. We now verify that this is the case. To do so, we evaluate a modified version of zero-shot prompting adapted to work for masked language models when we use no [MASK] tokens in the input. Specifically, on an input $x^+$ or $x^-$, we (1) take the logit vector corresponding to the last input token, (2) select the corresponding logit for ``positive'' and the corresponding logit for ``negative'' (the two candidate labels), (3) take the difference between these, resulting in $l_{pos}^+ - l_{neg}^+$ for $x^+$ and $l_{pos}^- - l_{neg}^-$ for $x^-$, respectively, then (4) take the difference between these, $(l_{pos}^+ - l_{neg}^+) - (l_{pos}^- - l_{neg}^-)$, to form the effective logit. Intuitively, this final expression should be large if the example is positive, and small if the example is negative. As before, we calibrate this zero-shot method so that its predictions are balanced.

We evaluate this method and find that it recovers non-trivial knowledge in the model outputs, but that it still performs substantially worse than CCS: calibrated zero-shot accuracy in this setting is $71.6\%$, compared to almost $94\%$ for CCS. 
This suggests that model outputs in this setting indeed aren't very useful, especially relative to CCS.

Overall, our results in this section show that CCS can still work in at least some settings where model outputs don't seem to be very useful. This provides additional evidence that out approach does not simply recover knowledge represented in the model outputs.

\section{Complete Sample Complexity Results}
\label{append:complete_sample_results}
\begin{figure*}[h]
\begin{center}
\includegraphics[width=0.6\textwidth]{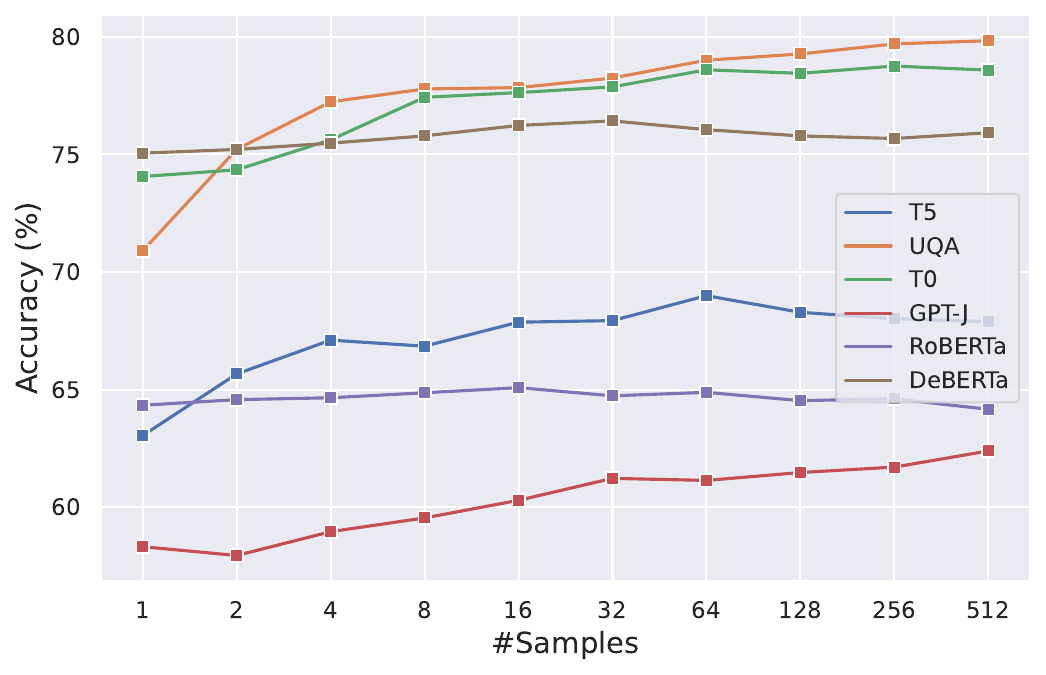}
\end{center}
\caption{
This figure shares the same setting with \Cref{fig:numdata}, but it contains all six models we consider.
}
\label{fig:numdata_all}
\end{figure*}

\begin{figure*}[h]
\begin{center}
\includegraphics[width=\textwidth]{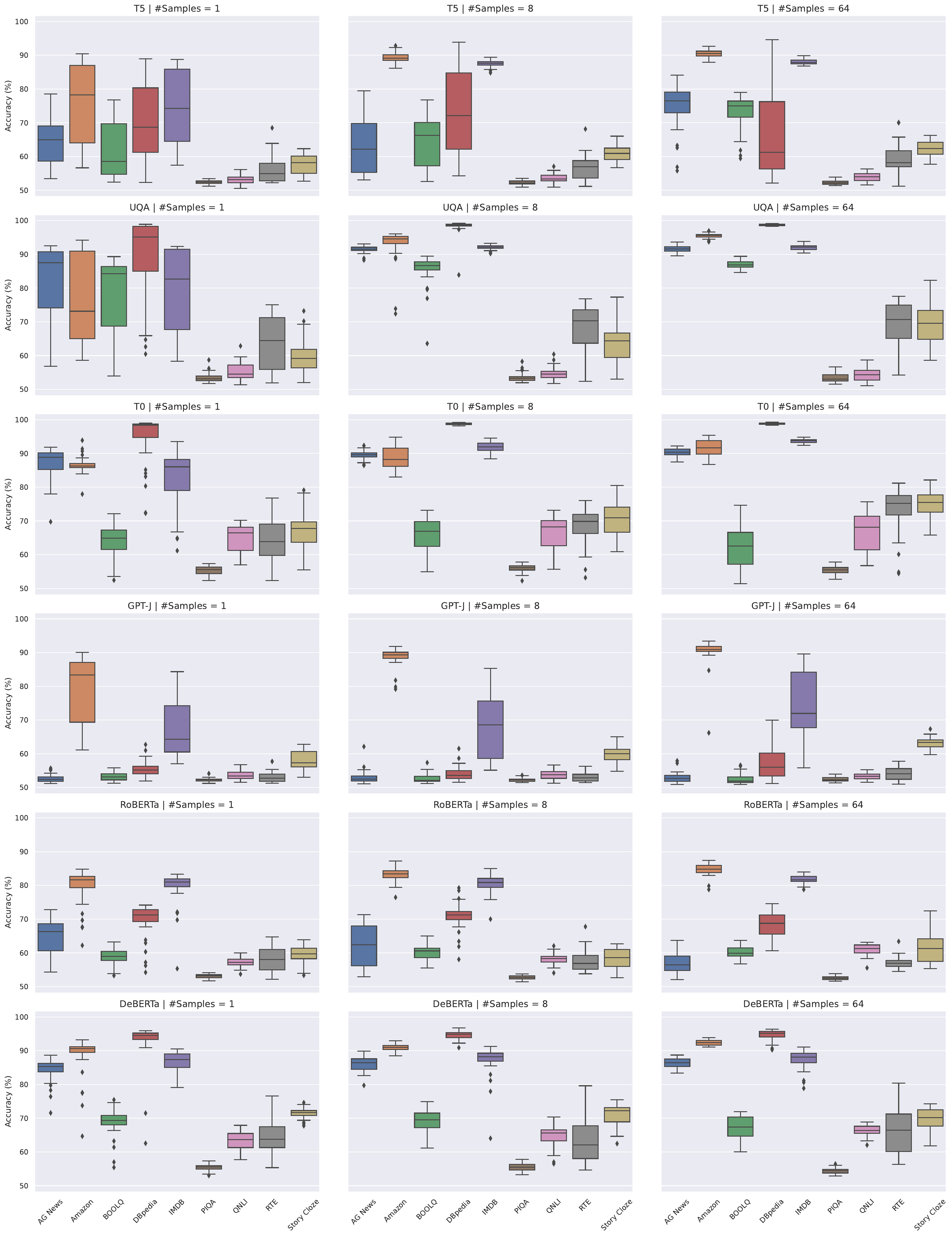}
\end{center}
\caption{
CCS performance when we only use $1,8,64$ examples, for all models and datasets. X-axis represents the dataset and y-axis is the accuracy in percentage. For each dataset, we select the prompt with highest 0-shot performance, and then randomly select $1,8,64$ data points from this prompt. We then perform CCS and test on all prompts of this dataset, where each value in the barplot corresponds to one prompt.
}
\label{fig:numdata_set-level}
\end{figure*}

In this section we show additional results on how the number of examples affects CCS. 
We show the performance (averaged across all datasets) for \textit{all} models in \Cref{fig:numdata_all}. 
We next show a more fine-grained results. 
Specifically, we select $\#\text{Samples} = 1,8,64$ and show dataset-level results in \Cref{fig:numdata_set-level} for each model. 

\clearpage
\section{Complete Transfer Results}
\label{sec:appx:transfer}
\begin{figure*}[h]
\begin{center}
\includegraphics[width=\textwidth]{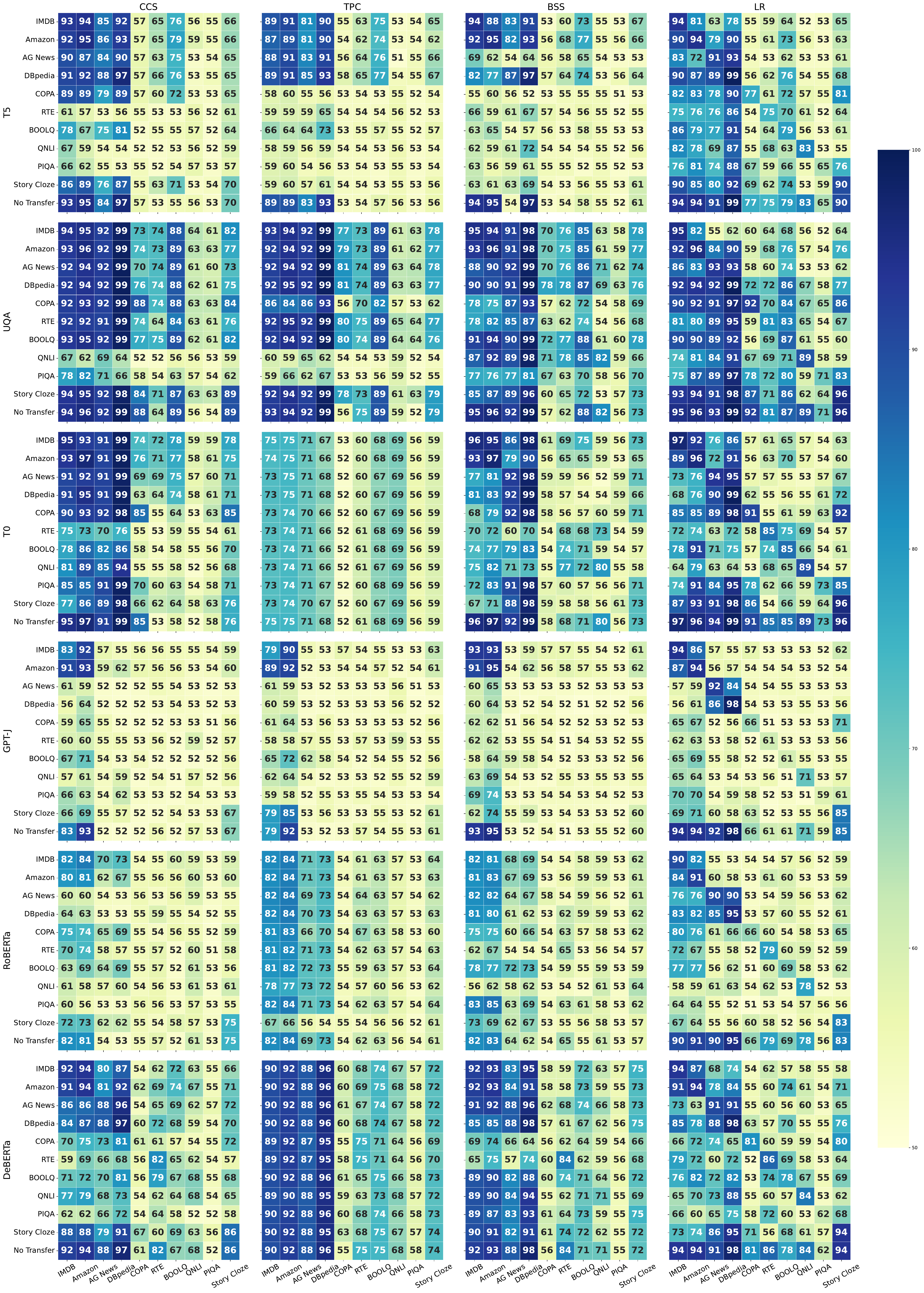}
\end{center}
\caption{
Transfer Accuracy for all models using CCS, BSS, TPC and LR. 
}
\label{fig:transfer_all}
\end{figure*}

Complete transfer results for CCS, TPC and BSS in all models are shown in \cref{fig:transfer_all}. Note that we normalize hidden states separately for each dataset when assessing transfer.

\clearpage
\section{Complete Intermediate Representations Results}
\label{append:sec:complete_intermediate}
\begin{figure*}[h!]
\begin{center}
\includegraphics[width=0.9\textwidth]{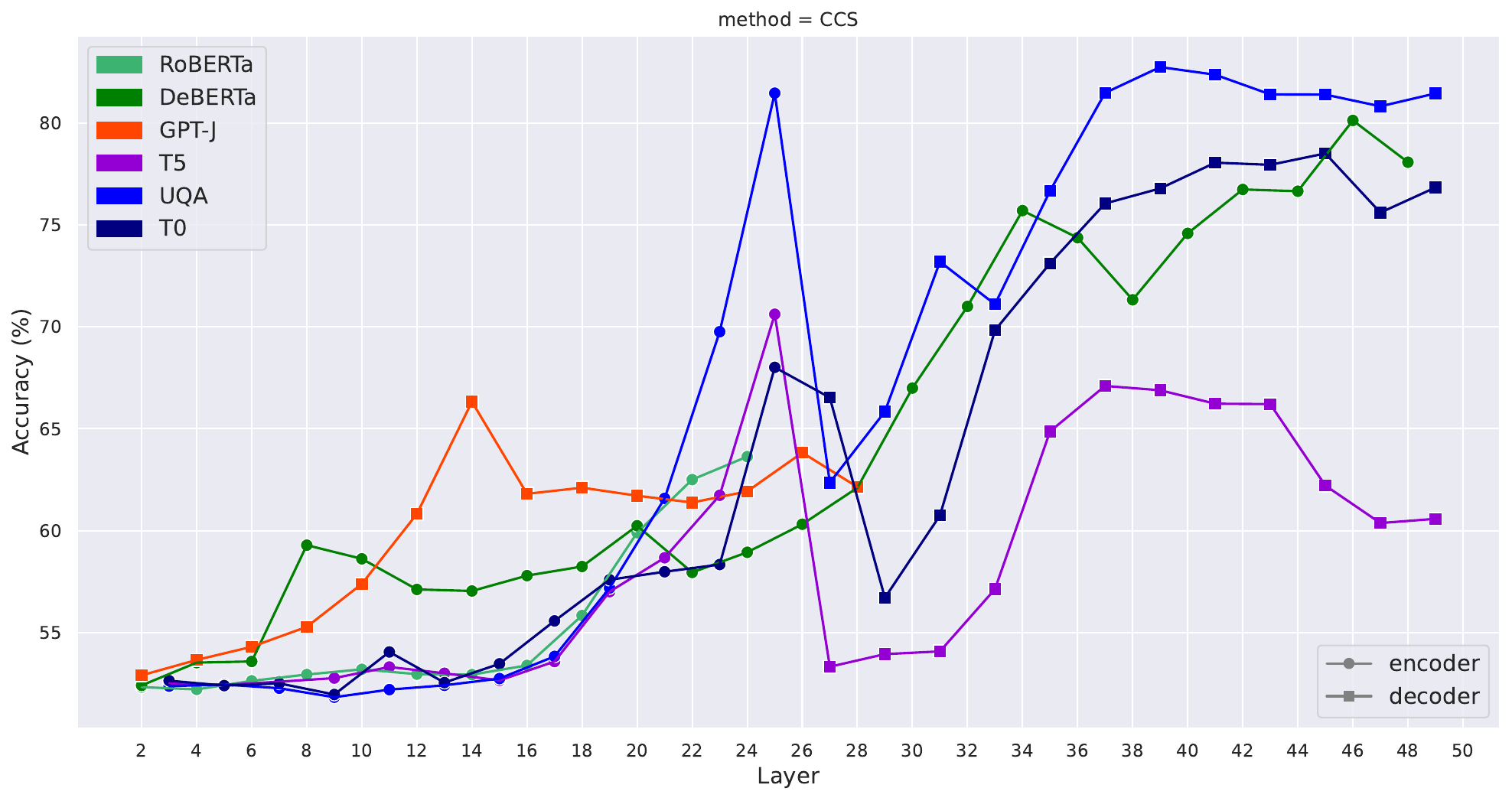}
\end{center}
\caption{
CCS performance when using the hidden states across different layers, using all six models we consider in the paper. 
}
\label{fig:crosslayer-CCS-6model}
\end{figure*}

\begin{figure*}[h!]
\begin{center}
\includegraphics[width=0.9\textwidth]{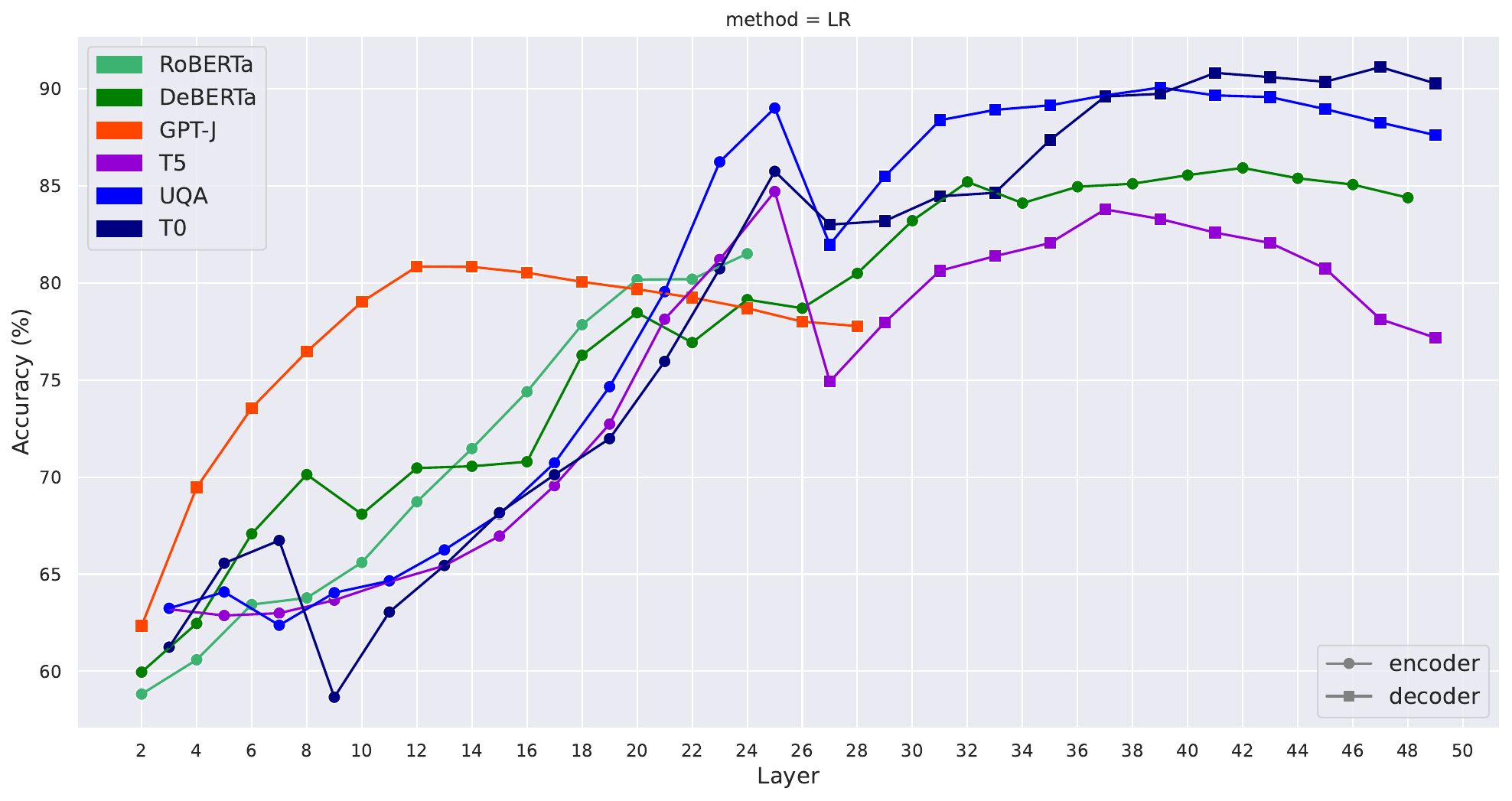}
\end{center}
\caption{
Linear regression performance when using the hidden states across different layers, using all six models we consider in the paper. This is the ceiling of all possible methods, and is supervised. 
}
\label{fig:crosslayer-LR-6model}
\end{figure*}

In this section we the performance of CCS and LR across all hidden layers in all six models. 
Specifically, for each model, we generate the hidden states for each dataset every 2 layers  (for both encoders and decoders).
Then, for each set of hidden states, we perform CCS and LR on each dataset separately, and average the performance across all datasets. 
We show CCS results in \Cref{fig:crosslayer-CCS-6model} and LR results in \Cref{fig:crosslayer-LR-6model}.

\clearpage
\section{CCS and CRC Implementation Details}
\label{sec:appx:implementation}
In this section, we provide further implementation details for CCS and CRC.

\subsection{Normalization}
\label{appendix:normalization}
As described in \Cref{subsec:method_ccs}, we normalize the features $\tilde{\phi}(x_i^+) = \phi(x_i^+) - \mu^+$, where $\mu^+$ is the mean of $\{\phi(x_i^+)\}_{i=1}^n$ (and similarly for $\tilde{\phi}(x_i^-)$). In practice, we also normalize the scale of the features by also dividing by the average norm of $\{\phi(x_i^+)\}_{i=1}^n$ times $\sqrt{d}$ (and similarly for $\tilde{\phi}(x_i^-)$). However, this is less essential than mean normalization and there are likely many reasonable choices for scale normalization.

\subsection{Contrast Pairs}
\label{appendix:contrast_pairs}
We use the Huggingface library \citep{Wolf2019HuggingFacesTS} for all of our experiments.
We use the standard tokenizer for each model, and always take the hidden state corresponding to the last token in a given layer indexed by $\text{idx}$. 
We show Huggingface-style pseudocode for extracting hidden states from in \Cref{alg:contrast}.

For encoder-only and decoder-only models, we provide the full input, $x^+$ or $x^-$, which includes both the question \textit{and} the proposed answer, to the model; see \Cref{append:tokenization} for more formatting and tokenization details. For encoder-decoder models, our input format depends on whether we are taking the encoder hidden states or the decoder hidden states. When we take the decoder hidden states, we input the question to the encoder, and input the candidate answer to the decoder. In contrast, when we take the encoder hidden states of an encoder-decoder model, we provide the full input (including the proposed answer) to the encoder, and ignore the decoder (simply passing the empty string to the decoder and ignoring its results). This is necessary to ensure that the inputs to the encoder are not identical across contrast pairs.

\begin{algorithm}[h!]
    \caption{Pseudocode for Getting Contrast Features}
	\label{alg:contrast} 
    \renewcommand{\algorithmicrequire}{\textbf{Input:}}
	\renewcommand{\algorithmicensure}{\textbf{Output:}}
	\begin{algorithmic}
		\REQUIRE Contrast Pairs Set $S $, model name $\text{mdl}$, layer index $\text{idx}$
        \STATE $\mathbf m = \text{transformers.AutoModel}(\text{mdl})$
        \STATE $\mathbf t = \text{transformers.AutoTokenizer}(\text{mdl})$
        \STATE $\mathcal C^+, C^- = [], []$
        \FOR{$(x^+, x^-) \textbf{ in } S$}
            \STATE token$^+$, token$^- = \mathbf t$.encode$(x^+)$, $\mathbf t$.encode$(x^-)$
            \STATE $\phi^+ = m(\text{token}^+, \text{output\_hiddenstates = True})\text{[``hidden\_states'']}[\text{idx}][-1]$        
            \STATE $\phi^- = m(\text{token}^-, \text{output\_hiddenstates = True})\text{[``hidden\_states'']}[\text{idx}][-1]$        
            \STATE $\mathcal C^+\text{.append}(\phi^+), \mathcal C^-\text{.append}(\phi^-)$
        \ENDFOR
        
		\ENSURE $[\mathcal C^+, \mathcal C^-]$ 
	\end{algorithmic} 
\end{algorithm}

\subsection{CCS}
\label{appendix:CCS}

Given contrast features from \Cref{alg:contrast}, CCS works by learning parameters $\theta$ and $b$ by minimizing $L_{CCS}(\theta, b)$ as defined in \Cref{subsec:method_ccs}.

In practice, we implement the bias $b$ by appending an additional dimension of $1$s features to the input, increasing it and $\theta$ to have dimension $d+1$.%
We randomly initialize $\theta$ to have unit norm. 
We then optimize the CCS loss $T=10$ times and select the run with lowest unsupervised loss. 
In practice, we train each time for $E=1000$ epochs with a learning rate $\eta = 0.01$ (which we found was good for consistently achieving low unsupervised loss) in each run.
Because we are only learning a linear probe on top of the features, training can be fast.

\subsection{CRC: Top Principal Component}
\label{appendix:CRC}

Intuitively, a direction that clusters examples well should have high variance. 
Motivated by this, a simple method is to cluster according to the Top Principal Component (TPC). 
We do this by first constructing contrast features from the normalized contrast pair activations, $\CC = \{\tilde{\phi}(x_i^+) - \tilde{\phi}(x_i^-)\}_{i=1}^n$, then projecting these examples on their top principal component using PCA.
We then treat examples that are less than $0$ as one cluster and examples that are greater than $0$ as the other cluster. 
We find that despite its simplicity this method can often also find truth-like features in model activations.
This method is similar in spirit to \citet{Bolukbasi2016ManIT}, except that our method doesn't require any labels; we leverage the fact that truth is consistent to construct contrast pairs in a purely unsupervised way.

\subsection{CRC: Bimodal Salience Search}

Bimodal Salience Search (BSS) is another variation of CRC.
One drawback of TPC is that variance is not an intrinsic quantity to a network's behavior, since different directions in representation space can be scaled without changing the behavior of the network as long as subsequent layers are rescaled accordingly. 
This motivates using a method that clusters examples regardless of the scale of different directions.

If examples are well-clustered, then the intra-cluster variance should be low while the inter-cluster variance (or total variance) should be high. 
Specifically, if we center a set of datapoints so that $0$ delineates the cluster boundary, 
then the points on either side of $0$ should have low variance compared to the points overall.
This suggests minimizing the following loss:
\begin{align}\label{eq:loss}
    L(\theta) = \frac{\var\{\theta^T c_i | \theta^T c_i < 0\} + \var\{\theta^T c_i | \theta^T c_i \geq 0\}}{\var\{\theta^T c_i\}}
\end{align}
where $c_i := \tilde{\phi}(x_i^+) - \tilde{\phi}(x_i^-) $ and where we use $\var\{z_i\}$ as shorthand to denote the variance of a set $\{z_i\}_{i=1}^n \subset \mathbb{R}$.
Because this objective is written in terms of a ratio, it is invariant to the overall scale of the direction, fixing that drawback of TPC. 
This loss is similar to that of Linear Discriminant Analysis (LDA) \citep{Fisher1936THEUO}, but unlike LDA is completely unsupervised.

In practice, we use an SGD-based optimizer to find a local optimum of $L(\theta)$. 
After computing the contrast features, we repeat the following process $T$ times.
We first initialize a random direction $\theta$ with unit norm.
Then for $E$ epochs, we calculate the loss according to \cref{eq:loss}, and update the direction $\theta$ by via projected gradient descent, each time projecting $\theta$ back onto the unit sphere. 
Finally, we select the direction $\theta$ that has the lowest loss among the $T$ directions we found, and predict based on this direction.
When we optimize the direction with multiple datasets, i.e. $S_1, ..., S_n$, we use the same algorithm, but average the loss across all datasets.
In practice we use $T = E = 20$, and use Adam optimizer \citep{kingma2014adam} with a learning rate of $0.1$.

\section{Statistical Significance}
\label{appx:sec:stat_significance}
Our main accuracy results for CCS and other methods (e.g. in \Cref{tab:avg} and elsewhere) are computed by evaluating the method on $40\%$ of the $1000$ (or $500$ in the case of COPA) examples sampled for each dataset, then averaging the resulting accuracy across $9$ prompts per dataset (on average), $10$ different datasets, and up to $5$ models. This corresponds to about $180$k samples in total; we performed subsampling in this way for computational efficiency reasons, as this is already substantial. For results where we average across datasets (i.e. most results in this paper), we average across $3800$ IID samples (the original examples sampled from each dataset), then usually also averaged across a very large number of correlated samples: ~$9$ prompts and (often) $5$ different models for a given sample.

We can compute an upper bound on the standard error of accuracies computed in this way by ignoring the averaging across different prompts and models: this gives us a simple but coarse upper bound of the standard error of $\frac{1}{2\sqrt{3800}} \approx 0.8\%$. Applying a Wald test, the Wald statistic is $W = \frac{\mu_0 - \hat{\mu}}{se(\hat{\mu})^2} \geq \frac{\mu_0 - \hat{\mu}}{(0.008)^2}$, where (for example) $\mu_0$ is zero-shot accuracy and $\hat{\mu}$ is CCS accuracy. $W$ is then $\chi^2$-distributed with one degree of freedom.

We find that our main claims comparing accuracies are statistically significant at a 0.00001 level. This includes the claims that CCS outperforms zero-shot, with accuracies of 71.2\% vs 67.2\%, respectively; that CCS is robust to misleading prompts while zero-shot isn’t in the setting we test, with accuracies of 83.8\% vs 70.9\%, respectively; that CCS on MLM-pretrained DeBERTa substantially outperforms DeBERTa zero-shot, with accuracies of 93.7\% vs 71.6\%, respectively; and so on.
Some minor observations are not necessarily statistically significant, such as that CCS (All Data) outperforms CCS on average (\Cref{subsec:ccs_outperforms}), but none of these are important for our main claims, and more powerful but complicated statistical tests would result in smaller p-values.

\clearpage
\section{Dataset Setup}
\label{sec:appx:setup}
In this section, we describe the setup of our data and prompts in detail.
We introduce all datasets we use and the way we convert them into a binary classification task.
Our prompts derive from \citep{Sanh2021MultitaskPT}\footnote{\url{https://github.com/bigscience-workshop/promptsource}}.

\textbf{Contrast Pair Example.} To illustrate how we construct contrast pairs, suppose we have a movie review ``[text] = I loved this movie.'' and the sentiment of this review belongs to ``[label0] = positive'' or  ``[label1] = negative''.
We first format it into a binary question-answering or classification question: ``[text] Is the sentiment of this example [label0] or [label1]? '' using an existing zero-shot prompt.
We then concatenate the question and candidate labels to create the contrast pairs:
\begin{align*}
x^+ &= \text{[prefix] Q: Is the sentiment of  ``[text]'' [label0] or [label1]? A: [label0]}\\
x^- &= \text{[prefix] Q: Is the sentiment of ``[text]'' [label0] or [label1]? A: [label1]}
\end{align*}
For instance, in this example, these would be:
\begin{align*}
x^+ &= \text{Q: Is the sentiment of ``I loved this movie.'' positive or negative? A: positive}\\
x^- &= \text{Q: Is the sentiment of ``I loved this movie.'' positive or negative? A: negative}
\end{align*}

\subsection{Tokenization}
\label{append:tokenization}
For all prompts we consider in this paper, we concatenate labels to the end of questions. 
The precise we do this depends on the model type. 
For encoder-decoder models such as T5, we replace line breaks (\verb|\n|) with spaces, and we add a space between the question and proposed answer if the last token of questions is not already a space.
For autoregressive models, we add a line break between the question and the proposed label if the last token of the question is neither a space nor a line break.
For DeBERTa, we add a \verb|[SEP]| split-token between the question and answer, and for RoBERTa, we add a \verb|</s><s>| between the question and answer.

Finally, we add the prefix (either prefix in \cref{fig:prefix}, or the empty string for the regular zero-shot setting) to the beginning of the prompt, then compute the hidden representations for this pair to obtain the contrast features $\mathcal C$. 

\subsection{Datasets}
\label{append:dataset}
We use ten datasets in our experiments.
For each dataset, we select 1000 data points (except COPA, which has only $500$ data points, so we only use those $500$). 
When possible, we use just the test / validation split, but in other cases this is not enough data so we also use examples from the train split.
For the most part this doesn't make a difference for our purposes because our methods are purely unsupervised.

It is important to note that T0 is an exception among the models, because but was trained on 9 out of the 10 datasets we consider (all except RTE). This may account for its anomalous performance.
UnifiedQA was trained on BoolQ, which is one of the datasets we test on, but was not trained on any other evaluation set we consider, which we don't expect has an important qualitative effect.

We now enumerate the prompts we use for each dataset.

\subsubsection{AG\_News}
AG\_News is a topic classification dataset from \citep{zhang2015character}. %
Given a news paragraph, the label corresponds to the topic of this news, including ``World politics'', ``Sports'', ``Business'' and ``Science and technology''.
For each example, we randomly select another label (which is incorrect), and randomly assign the correct label and this incorrect label to ``label0'' and ``label1''. With this, we construct the classification task using the following 8 prompts:
\begin{enumerate}
\item [1]
[prefix]Consider the following example: ``` [text] '''
Choice 1: [label0]. Choice 2: [label1].Between choice 1 and choice 2, the topic of this example is [label]

\item [2]
[prefix]Consider the following example: ``` [text] '''
Choice 1: [label0]. Choice 2: [label1].Between choice 1 and choice 2, what is the topic of this example?
[label]

\item [3]
[prefix]Consider the following example: ``` [text] '''
Which is the topic of this example, choice 1: [label0], or choice 2: [label1]?
[label]

\item [4]
[prefix][text]
Choice 1: [label0]. Choice 2: [label1].Between choice 1 and choice 2, the topic of this example is [label]

\item [5]
[prefix][text]
Choice 1: [label0]. Choice 2: [label1].Between choice 1 and choice 2, what is the topic of this example?
[label]

\item [6]
[prefix][text]
Which is the topic of this example, choice 1: [label0], or choice 2: [label1]?
[label]

\item [7]
[prefix][text]
What label best describes this news article, choice 1: [label0], or choice 2: [label1]?
[label]

\item [8]
[prefix][text]
Which section of a newspaper would this article likely appear in, choice 1: [label0], or choice 2: [label1]?
[label]
\end{enumerate}

Here the last ``[label]'' is ``choice 1'' for $x^+$ and ``choice 2'' for $x^-$. Notice that we only specify ``[label]'' for the prompts that are constructed manually. For prompts in \citep{Sanh2021MultitaskPT}, we leave their labels unchanged. (For example, we use ``choice 1'' or ``choice 2'' but they can use ``Animal'' or ``Plant'')

\subsubsection{Amazon\_polarity}
Amazon\_polarity is a sentiment classification task from \citep{mcauley2013hidden}.
The content is the review of goods in Amazon, and the label can be ``negative'' or ``positive''.
We use 11 different prompts in this dataset. We first take all prompts from \citep{Sanh2021MultitaskPT} (Page 164, 9 prompts in total), and the two of our own as follows:

\begin{enumerate}
    \item [1] [prefix]Consider the following example: ``` [content] '''
Between [label0] and [label1], the sentiment of this example is
[label]

    \item [2] [prefix]Consider the following example: ``` [content] '''
    Between [label0] and [label1], which is the sentiment of this example?
    [label]
\end{enumerate}

Here ``[label]'' is ``negative'' for $x^+$ and ``positive'' for $x^-$.

\subsubsection{BOOLQ}
BOOLQ is a QA task where each example consists a yes/no question from \citep{clark2019boolq}. We directly use the 10 prompts from \citep{Sanh2021MultitaskPT} (Page 146)

\subsubsection{COPA}
COPA is a causal reasoning task to determine either the cause or the effect of a given premise \citep{roemmele2011choice}. 
Here the label is a short sentence. 
We use 10 prompts, where 9 are from \citep{Sanh2021MultitaskPT} (Page 177), and we add one more prompt:

\begin{enumerate}
\item [1] [prefix]Consider the following premise: ``` [premise] ''' Choice 1: [choice1]
Choice 2: [choice2]
Q: Which one is more likely to be the [question], choice 1 or choice 2?
[label]

\end{enumerate}

\subsubsection{DBpedia\_14}
DBpedia\_14 is a topic classification dataset constructed by picking 14 non-overlapping classes from DBpedia 2014 \citep{lehmann2015dbpedia}. We manually create 8 prompts. For each example, we randomly select the incorrect label from the remaining 13 classes, and randomly assign the correct label and this incorrect label to ``[label0]'' and ``[label1]''.

\begin{enumerate}
\item [1] [prefix]Consider the following example: ``` [content] '''
Choice 1: [label0]. Choice 2: [label1].Between choice 1 and choice 2, the topic of this example is [label]

\item [2] [prefix]Consider the following example: ``` [content] '''
Choice 1: [label0]. Choice 2: [label1].Between choice 1 and choice 2, what is the topic of this example?
[label]

\item [3] [prefix]Consider the following example: ``` [content] '''
Which is the topic of this example, choice 1: [label0], or choice 2: [label1]?
[label]

\item [4] [prefix][content]
Choice 1: [label0]. Choice 2: [label1].Between choice 1 and choice 2, the topic of this example is [label]

\item [5] [prefix][content]
Choice 1: [label0]. Choice 2: [label1].Between choice 1 and choice 2, what is the topic of this example?
[label]

\item [6] [prefix][content]
Which is the topic of this example, choice 1: [label0], or choice 2: [label1]?
[label]

\item [7] [prefix][content]
What category does the paragraph belong to, choice 1: [label0], or choice 2: [label1]?
[label]

\item [8] [prefix][content]
What label best describes this paragraph, choice 1: [label0], or choice 2: [label1]?
[label]
\end{enumerate}

Here ``[label]'' is ``choice 1'' for $x^+$ and ``choice 2'' for $x^-$.

\subsubsection{IMDB}
IMDB is a sentiment dataset from \citep{maas2011learning}. Given a movie review, the label is either ``[label0] = negative'' or ``[label1] = positive. We use 13 prompts, where 11 are from \citep{Sanh2021MultitaskPT} (Page 168), and the rest two are as follows:

\begin{enumerate}
 \item [1] [prefix]Consider the following example: ''' [text] '''
Between [label0] and [label1], the sentiment of this example is
[label]

\item [2] [prefix]Consider the following example: ''' [text] '''
Between [label0] and [label1], which is the sentiment of this example?
[label]
\end{enumerate}

Here ``[label]'' is ``negative'' for $x^+$ and ``positive'' for $x^-$.

\subsubsection{PIQA}
The PIQA dataset measures the physical commonsense reasoning ability of models. We use 11 prompts for PIQA, all of which are from \citep{Sanh2021MultitaskPT}(Page 160). 
The label is a complete sentence that can be the solution of the question.

\subsubsection{QNLI}
QNLI from \citep{rajpurkar2016squad} is a question-answering dataset consisting of question-paragraph pairs, where one of the sentences in the paragraph (drawn from Wikipedia) contains the answer to the corresponding question (written by an annotator). 
We use 5 prompts from \citep{Sanh2021MultitaskPT}. 
The label is either ``yes'' or ``no'' depending on whether the information in the paragraph is enough to answer the paragraph.

\subsubsection{RTE}
RTE is a textual entailment dataset \citep{wang2018glue}. 
The label corresponds to whether the text entails the hypotheses. 
We use 11 prompts, where 10 are from \citep{Sanh2021MultitaskPT}(Page 48), and where we manually add one more prompt:

\begin{enumerate}
\item [1] [prefix][premise]
Question: Does this imply that "[hypothesis]", yes or no?
[label]
\end{enumerate}

Here ``[label]'' is ``yes'' for $x^+$ and ``no'' for $x^-$.

\subsubsection{Story\_Cloze}
Story Cloze is a story completion task from \citep{mostafazadeh2017lsdsem}. 
Given a short story, the task is to determine which of two endings is more likely to continue that story. 
We use 9 prompts, where 6 are from \citep{Sanh2021MultitaskPT}, and the remaining 3 are as follows:

\begin{enumerate}
\item [1] [prefix]Consider the following story: ``` [input\_sentence\_1] [input\_sentence\_2] [input\_sentence\_3] [input\_sentence\_4] '''
Choice 1: [sentence\_quiz1]
Choice 2: [sentence\_quiz2]
Which is the more plausible ending of this story, choice 1 or choice 2?
[label]

\item [2] [prefix]Consider the following story: ``` [input\_sentence\_1] [input\_sentence\_2] [input\_sentence\_3] [input\_sentence\_4] '''
Choice 1: [sentence\_quiz1]
Choice 2: [sentence\_quiz2]
Which is the more plausible ending of this story?
[label]

\item [3] [prefix][input\_sentence\_1] [input\_sentence\_2] [input\_sentence\_3] [input\_sentence\_4]
Choice 1: [sentence\_quiz1]
Choice 2: [sentence\_quiz2]
Which is the more plausible ending of this story, choice 1 or choice 2?
[label]

\end{enumerate}
Here ``[label]'' is ``choice 1'' for $x^+$ and ``choice 2'' for $x^-$.

%% file: main.bbl
\begin{thebibliography}{58}
\providecommand{\natexlab}[1]{#1}
\providecommand{\url}[1]{\texttt{#1}}
\expandafter\ifx\csname urlstyle\endcsname\relax
  \providecommand{\doi}[1]{doi: #1}\else
  \providecommand{\doi}{doi: \begingroup \urlstyle{rm}\Url}\fi

\bibitem[Askell et~al.(2021)Askell, Bai, Chen, Drain, Ganguli, Henighan, Jones,
  Joseph, Mann, DasSarma, Elhage, Hatfield-Dodds, Hernandez, Kernion, Ndousse,
  Olsson, Amodei, Brown, Clark, McCandlish, Olah, and Kaplan]{Askell2021AGL}
Amanda Askell, Yushi Bai, Anna Chen, Dawn Drain, Deep Ganguli, T.~J. Henighan,
  Andy Jones, Nicholas Joseph, Benjamin Mann, Nova DasSarma, Nelson Elhage, Zac
  Hatfield-Dodds, Danny Hernandez, John Kernion, Kamal Ndousse, Catherine
  Olsson, Dario Amodei, Tom~B. Brown, Jack Clark, Sam McCandlish, Christopher
  Olah, and Jared Kaplan.
\newblock A general language assistant as a laboratory for alignment.
\newblock \emph{ArXiv}, abs/2112.00861, 2021.

\bibitem[Bai et~al.(2022)Bai, Jones, Ndousse, Askell, Chen, DasSarma, Drain,
  Fort, Ganguli, Henighan, Joseph, Kadavath, Kernion, Conerly, El-Showk,
  Elhage, Hatfield-Dodds, Hernandez, Hume, Johnston, Kravec, Lovitt, Nanda,
  Olsson, Amodei, Brown, Clark, McCandlish, Olah, Mann, and
  Kaplan]{Bai2022TrainingAH}
Yushi Bai, Andy Jones, Kamal Ndousse, Amanda Askell, Anna Chen, Nova DasSarma,
  Dawn Drain, Stanislav Fort, Deep Ganguli, T.~J. Henighan, Nicholas Joseph,
  Saurav Kadavath, John Kernion, Tom Conerly, Sheer El-Showk, Nelson Elhage,
  Zac Hatfield-Dodds, Danny Hernandez, Tristan Hume, Scott Johnston, Shauna
  Kravec, Liane Lovitt, Neel Nanda, Catherine Olsson, Dario Amodei, Tom~B.
  Brown, Jack Clark, Sam McCandlish, Christopher Olah, Benjamin Mann, and Jared
  Kaplan.
\newblock Training a helpful and harmless assistant with reinforcement learning
  from human feedback.
\newblock \emph{ArXiv}, abs/2204.05862, 2022.

\bibitem[Beltagy et~al.(2022)Beltagy, Cohan, IV, Min, and
  Singh]{Beltagy2022ZeroAF}
Iz~Beltagy, Arman Cohan, Robert~Logan IV, Sewon Min, and Sameer Singh.
\newblock Zero- and few-shot nlp with pretrained language models.
\newblock In \emph{ACL}, 2022.

\bibitem[Bender et~al.(2021)Bender, Gebru, McMillan-Major, and
  Shmitchell]{Bender2021OnTD}
Emily~M. Bender, Timnit Gebru, Angelina McMillan-Major, and Shmargaret
  Shmitchell.
\newblock On the dangers of stochastic parrots: Can language models be too big?
\newblock \emph{Proceedings of the 2021 ACM Conference on Fairness,
  Accountability, and Transparency}, 2021.

\bibitem[Bisk et~al.(2020)Bisk, Zellers, Gao, Choi, et~al.]{bisk2020piqa}
Yonatan Bisk, Rowan Zellers, Jianfeng Gao, Yejin Choi, et~al.
\newblock Piqa: Reasoning about physical commonsense in natural language.
\newblock In \emph{Proceedings of the AAAI Conference on Artificial
  Intelligence}, volume~34, pp.\  7432--7439, 2020.

\bibitem[Bolukbasi et~al.(2016)Bolukbasi, Chang, Zou, Saligrama, and
  Kalai]{Bolukbasi2016ManIT}
Tolga Bolukbasi, Kai-Wei Chang, James~Y. Zou, Venkatesh Saligrama, and
  Adam~Tauman Kalai.
\newblock Man is to computer programmer as woman is to homemaker? debiasing
  word embeddings.
\newblock In \emph{NIPS}, 2016.

\bibitem[Bommasani et~al.(2021)Bommasani, Hudson, Adeli, Altman, Arora, von
  Arx, Bernstein, Bohg, Bosselut, Brunskill, Brynjolfsson, Buch, Card,
  Castellon, Chatterji, Chen, Creel, Davis, Demszky, Donahue, Doumbouya,
  Durmus, Ermon, Etchemendy, Ethayarajh, Fei-Fei, Finn, Gale, Gillespie, Goel,
  Goodman, Grossman, Guha, Hashimoto, Henderson, Hewitt, Ho, Hong, Hsu, Huang,
  Icard, Jain, Jurafsky, Kalluri, Karamcheti, Keeling, Khani, Khattab, Koh,
  Krass, Krishna, Kuditipudi, Kumar, Ladhak, Lee, Lee, Leskovec, Levent, Li,
  Li, Ma, Malik, Manning, Mirchandani, Mitchell, Munyikwa, Nair, Narayan,
  Narayanan, Newman, Nie, Niebles, Nilforoshan, Nyarko, Ogut, Orr,
  Papadimitriou, Park, Piech, Portelance, Potts, Raghunathan, Reich, Ren, Rong,
  Roohani, Ruiz, Ryan, R'e, Sadigh, Sagawa, Santhanam, Shih, Srinivasan,
  Tamkin, Taori, Thomas, Tram{\`e}r, Wang, Wang, Wu, Wu, Wu, Xie, Yasunaga,
  You, Zaharia, Zhang, Zhang, Zhang, Zhang, Zheng, Zhou, and
  Liang]{Bommasani2021OnTO}
Rishi Bommasani, Drew~A. Hudson, Ehsan Adeli, Russ Altman, Simran Arora, Sydney
  von Arx, Michael~S. Bernstein, Jeannette Bohg, Antoine Bosselut, Emma
  Brunskill, Erik Brynjolfsson, S.~Buch, Dallas Card, Rodrigo Castellon,
  Niladri~S. Chatterji, Annie~S. Chen, Kathleen~A. Creel, Jared Davis, Dora
  Demszky, Chris Donahue, Moussa Doumbouya, Esin Durmus, Stefano Ermon, John
  Etchemendy, Kawin Ethayarajh, Li~Fei-Fei, Chelsea Finn, Trevor Gale,
  Lauren~E. Gillespie, Karan Goel, Noah~D. Goodman, Shelby Grossman, Neel Guha,
  Tatsunori Hashimoto, Peter Henderson, John Hewitt, Daniel~E. Ho, Jenny Hong,
  Kyle Hsu, Jing Huang, Thomas~F. Icard, Saahil Jain, Dan Jurafsky, Pratyusha
  Kalluri, Siddharth Karamcheti, Geoff Keeling, Fereshte Khani, O.~Khattab,
  Pang~Wei Koh, Mark~S. Krass, Ranjay Krishna, Rohith Kuditipudi, Ananya Kumar,
  Faisal Ladhak, Mina Lee, Tony Lee, Jure Leskovec, Isabelle Levent, Xiang~Lisa
  Li, Xuechen Li, Tengyu Ma, Ali Malik, Christopher~D. Manning, Suvir~P.
  Mirchandani, Eric Mitchell, Zanele Munyikwa, Suraj Nair, Avanika Narayan,
  Deepak Narayanan, Benjamin Newman, Allen Nie, Juan~Carlos Niebles, Hamed
  Nilforoshan, J.~F. Nyarko, Giray Ogut, Laurel Orr, Isabel Papadimitriou,
  Joon~Sung Park, Chris Piech, Eva Portelance, Christopher Potts, Aditi
  Raghunathan, Robert Reich, Hongyu Ren, Frieda Rong, Yusuf~H. Roohani, Camilo
  Ruiz, Jack Ryan, Christopher R'e, Dorsa Sadigh, Shiori Sagawa, Keshav
  Santhanam, Andy Shih, Krishna~Parasuram Srinivasan, Alex Tamkin, Rohan Taori,
  Armin~W. Thomas, Florian Tram{\`e}r, Rose~E. Wang, William Wang, Bohan Wu,
  Jiajun Wu, Yuhuai Wu, Sang~Michael Xie, Michihiro Yasunaga, Jiaxuan You,
  Matei~A. Zaharia, Michael Zhang, Tianyi Zhang, Xikun Zhang, Yuhui Zhang,
  Lucia Zheng, Kaitlyn Zhou, and Percy Liang.
\newblock On the opportunities and risks of foundation models.
\newblock \emph{ArXiv}, abs/2108.07258, 2021.

\bibitem[Brown et~al.(2020)Brown, Mann, Ryder, Subbiah, Kaplan, Dhariwal,
  Neelakantan, Shyam, Sastry, Askell, Agarwal, Herbert-Voss, Krueger, Henighan,
  Child, Ramesh, Ziegler, Wu, Winter, Hesse, Chen, Sigler, Litwin, Gray, Chess,
  Clark, Berner, McCandlish, Radford, Sutskever, and
  Amodei]{Brown2020LanguageMA}
Tom~B. Brown, Benjamin Mann, Nick Ryder, Melanie Subbiah, Jared Kaplan,
  Prafulla Dhariwal, Arvind Neelakantan, Pranav Shyam, Girish Sastry, Amanda
  Askell, Sandhini Agarwal, Ariel Herbert-Voss, Gretchen Krueger, T.~J.
  Henighan, Rewon Child, Aditya Ramesh, Daniel~M. Ziegler, Jeff Wu, Clemens
  Winter, Christopher Hesse, Mark Chen, Eric Sigler, Mateusz Litwin, Scott
  Gray, Benjamin Chess, Jack Clark, Christopher Berner, Sam McCandlish, Alec
  Radford, Ilya Sutskever, and Dario Amodei.
\newblock Language models are few-shot learners.
\newblock \emph{ArXiv}, abs/2005.14165, 2020.

\bibitem[Christiano et~al.(2022)Christiano, Cotra, and Xu]{ELK}
Paul Christiano, Ajeya Cotra, and Mark Xu.
\newblock Eliciting latent knowledge.
\newblock Technical report, ARC, 2022.
\newblock URL
  \url{https://docs.google.com/document/d/1WwsnJQstPq91_Yh-Ch2XRL8H_EpsnjrC1dwZXR37PC8/edit}.

\bibitem[Christiano et~al.(2017)Christiano, Leike, Brown, Martic, Legg, and
  Amodei]{Christiano2017DeepRL}
Paul~Francis Christiano, Jan Leike, Tom~B. Brown, Miljan Martic, Shane Legg,
  and Dario Amodei.
\newblock Deep reinforcement learning from human preferences.
\newblock \emph{ArXiv}, abs/1706.03741, 2017.

\bibitem[Christiano et~al.(2018)Christiano, Shlegeris, and
  Amodei]{Christiano2018SupervisingSL}
Paul~Francis Christiano, Buck Shlegeris, and Dario Amodei.
\newblock Supervising strong learners by amplifying weak experts.
\newblock \emph{ArXiv}, abs/1810.08575, 2018.

\bibitem[Clark et~al.(2019)Clark, Lee, Chang, Kwiatkowski, Collins, and
  Toutanova]{clark2019boolq}
Christopher Clark, Kenton Lee, Ming-Wei Chang, Tom Kwiatkowski, Michael
  Collins, and Kristina Toutanova.
\newblock Boolq: Exploring the surprising difficulty of natural yes/no
  questions.
\newblock \emph{arXiv preprint arXiv:1905.10044}, 2019.

\bibitem[Evans et~al.(2021)Evans, Cotton-Barratt, Finnveden, Bales, Balwit,
  Wills, Righetti, and Saunders]{Evans2021TruthfulAD}
Owain Evans, Owen Cotton-Barratt, Lukas Finnveden, Adam Bales, Avital Balwit,
  Peter Wills, Luca Righetti, and William Saunders.
\newblock Truthful ai: Developing and governing ai that does not lie.
\newblock \emph{ArXiv}, abs/2110.06674, 2021.

\bibitem[FAIR et~al.(2022)FAIR, Bakhtin, Brown, Dinan, Farina, Flaherty, Fried,
  Goff, Gray, Hu, Jacob, Komeili, Konath, Kwon, Lerer, Lewis, Miller, Mitts,
  Renduchintala, Roller, Rowe, Shi, Spisak, Wei, Wu, Zhang, and
  Zijlstra]{Bakhtin2022HumanlevelPI}
Meta FAIR, Anton Bakhtin, Noam Brown, Emily Dinan, Gabriele Farina, Colin
  Flaherty, Daniel Fried, Andrew Goff, Jonathan Gray, Hengyuan Hu, Athul~Paul
  Jacob, Mojtaba Komeili, Karthik Konath, Minae Kwon, Adam Lerer, Mike Lewis,
  Alexander~H. Miller, Sasha Mitts, Adithya Renduchintala, Stephen Roller, Dirk
  Rowe, Weiyan Shi, Joe Spisak, Alexander Wei, David Wu, Hugh Zhang, and Markus
  Zijlstra.
\newblock Human-level play in the game of diplomacy by combining language
  models with strategic reasoning.
\newblock \emph{Science}, pp.\  eade9097, 2022.

\bibitem[Fisher(1936)]{Fisher1936THEUO}
Rory~A. Fisher.
\newblock The use of multiple measurements in taxonomic problems.
\newblock \emph{Annals of Human Genetics}, 7:\penalty0 179--188, 1936.

\bibitem[He et~al.(2021)He, Liu, Gao, and Chen]{He2021DeBERTaDB}
Pengcheng He, Xiaodong Liu, Jianfeng Gao, and Weizhu Chen.
\newblock Deberta: Decoding-enhanced bert with disentangled attention.
\newblock \emph{ArXiv}, abs/2006.03654, 2021.

\bibitem[Hendrycks et~al.(2021)Hendrycks, Carlini, Schulman, and
  Steinhardt]{Hendrycks2021UnsolvedPI}
Dan Hendrycks, Nicholas Carlini, John Schulman, and Jacob Steinhardt.
\newblock Unsolved problems in ml safety.
\newblock \emph{ArXiv}, abs/2109.13916, 2021.

\bibitem[Irving et~al.(2018)Irving, Christiano, and Amodei]{Irving2018AISV}
Geoffrey Irving, Paul~Francis Christiano, and Dario Amodei.
\newblock Ai safety via debate.
\newblock \emph{ArXiv}, abs/1805.00899, 2018.

\bibitem[Jung et~al.(2022)Jung, Qin, Welleck, Brahman, Bhagavatula, Bras, and
  Choi]{Jung2022MaieuticPL}
Jaehun Jung, Lianhui Qin, Sean Welleck, Faeze Brahman, Chandra Bhagavatula,
  Ronan~Le Bras, and Yejin Choi.
\newblock Maieutic prompting: Logically consistent reasoning with recursive
  explanations.
\newblock \emph{ArXiv}, abs/2205.11822, 2022.

\bibitem[Kenton et~al.(2021)Kenton, Everitt, Weidinger, Gabriel, Mikulik, and
  Irving]{Kenton2021AlignmentOL}
Zachary Kenton, Tom Everitt, Laura Weidinger, Iason Gabriel, Vladimir Mikulik,
  and Geoffrey Irving.
\newblock Alignment of language agents.
\newblock \emph{ArXiv}, abs/2103.14659, 2021.

\bibitem[Khashabi et~al.(2020)Khashabi, Min, Khot, Sabharwal, Tafjord, Clark,
  and Hajishirzi]{Khashabi2020UnifiedQACF}
Daniel Khashabi, Sewon Min, Tushar Khot, Ashish Sabharwal, Oyvind Tafjord,
  Peter Clark, and Hannaneh Hajishirzi.
\newblock Unifiedqa: Crossing format boundaries with a single qa system.
\newblock In \emph{FINDINGS}, 2020.

\bibitem[Kim et~al.(2022)Kim, Kim, Cho, Jo, Lee, goo Lee, Yoo, and
  Kim]{Kim2022GroundTruthLM}
Junyeob Kim, Hyuhng~Joon Kim, Hyunsoo Cho, Hwiyeol Jo, Sang-Woo Lee, Sang goo
  Lee, Kang~Min Yoo, and Taeuk Kim.
\newblock Ground-truth labels matter: A deeper look into input-label
  demonstrations.
\newblock \emph{ArXiv}, abs/2205.12685, 2022.

\bibitem[Kingma \& Ba(2014)Kingma and Ba]{kingma2014adam}
Diederik~P Kingma and Jimmy Ba.
\newblock Adam: A method for stochastic optimization.
\newblock \emph{arXiv preprint arXiv:1412.6980}, 2014.

\bibitem[Lehmann et~al.(2015)Lehmann, Isele, Jakob, Jentzsch, Kontokostas,
  Mendes, Hellmann, Morsey, Van~Kleef, Auer, et~al.]{lehmann2015dbpedia}
Jens Lehmann, Robert Isele, Max Jakob, Anja Jentzsch, Dimitris Kontokostas,
  Pablo~N Mendes, Sebastian Hellmann, Mohamed Morsey, Patrick Van~Kleef,
  S{\"o}ren Auer, et~al.
\newblock Dbpedia--a large-scale, multilingual knowledge base extracted from
  wikipedia.
\newblock \emph{Semantic web}, 6\penalty0 (2):\penalty0 167--195, 2015.

\bibitem[Leike et~al.(2018)Leike, Krueger, Everitt, Martic, Maini, and
  Legg]{Leike2018ScalableAA}
Jan Leike, David Krueger, Tom Everitt, Miljan Martic, Vishal Maini, and Shane
  Legg.
\newblock Scalable agent alignment via reward modeling: a research direction.
\newblock \emph{ArXiv}, abs/1811.07871, 2018.

\bibitem[Lin et~al.(2022)Lin, Hilton, and Evans]{Lin2022TruthfulQAMH}
Stephanie~C. Lin, Jacob Hilton, and Owain Evans.
\newblock Truthfulqa: Measuring how models mimic human falsehoods.
\newblock In \emph{ACL}, 2022.

\bibitem[Liu et~al.(2022)Liu, Yuan, Fu, Jiang, Hayashi, and
  Neubig]{Liu2022PretrainPA}
Pengfei Liu, Weizhe Yuan, Jinlan Fu, Zhengbao Jiang, Hiroaki Hayashi, and
  Graham Neubig.
\newblock Pre-train, prompt, and predict: A systematic survey of prompting
  methods in natural language processing.
\newblock \emph{ACM Computing Surveys (CSUR)}, 2022.

\bibitem[Liu et~al.(2019)Liu, Ott, Goyal, Du, Joshi, Chen, Levy, Lewis,
  Zettlemoyer, and Stoyanov]{Liu2019RoBERTaAR}
Yinhan Liu, Myle Ott, Naman Goyal, Jingfei Du, Mandar Joshi, Danqi Chen, Omer
  Levy, Mike Lewis, Luke Zettlemoyer, and Veselin Stoyanov.
\newblock Roberta: A robustly optimized bert pretraining approach.
\newblock \emph{ArXiv}, abs/1907.11692, 2019.

\bibitem[Loshchilov \& Hutter(2017)Loshchilov and
  Hutter]{Loshchilov2017FixingWD}
Ilya Loshchilov and Frank Hutter.
\newblock Fixing weight decay regularization in adam.
\newblock \emph{ArXiv}, abs/1711.05101, 2017.

\bibitem[Lu et~al.(2022)Lu, Bartolo, Moore, Riedel, and
  Stenetorp]{Lu2022FantasticallyOP}
Yao Lu, Max Bartolo, Alastair Moore, Sebastian Riedel, and Pontus Stenetorp.
\newblock Fantastically ordered prompts and where to find them: Overcoming
  few-shot prompt order sensitivity.
\newblock In \emph{ACL}, 2022.

\bibitem[Maas et~al.(2011)Maas, Daly, Pham, Huang, Ng, and
  Potts]{maas2011learning}
Andrew Maas, Raymond~E Daly, Peter~T Pham, Dan Huang, Andrew~Y Ng, and
  Christopher Potts.
\newblock Learning word vectors for sentiment analysis.
\newblock In \emph{Proceedings of the 49th annual meeting of the association
  for computational linguistics: Human language technologies}, pp.\  142--150,
  2011.

\bibitem[Maynez et~al.(2020)Maynez, Narayan, Bohnet, and
  McDonald]{Maynez2020OnFA}
Joshua Maynez, Shashi Narayan, Bernd Bohnet, and Ryan~T. McDonald.
\newblock On faithfulness and factuality in abstractive summarization.
\newblock \emph{ArXiv}, abs/2005.00661, 2020.

\bibitem[McAuley \& Leskovec(2013)McAuley and Leskovec]{mcauley2013hidden}
Julian McAuley and Jure Leskovec.
\newblock Hidden factors and hidden topics: understanding rating dimensions
  with review text.
\newblock In \emph{Proceedings of the 7th ACM conference on Recommender
  systems}, pp.\  165--172, 2013.

\bibitem[Menick et~al.(2022)Menick, Trebacz, Mikulik, Aslanides, Song,
  Chadwick, Glaese, Young, Campbell-Gillingham, Irving, and
  McAleese]{Menick2022TeachingLM}
Jacob Menick, Maja Trebacz, Vladimir Mikulik, John Aslanides, Francis Song,
  Martin Chadwick, Mia Glaese, Susannah Young, Lucy Campbell-Gillingham,
  Geoffrey Irving, and Nathan McAleese.
\newblock Teaching language models to support answers with verified quotes.
\newblock \emph{ArXiv}, abs/2203.11147, 2022.

\bibitem[Min et~al.(2022{\natexlab{a}})Min, Lewis, Zettlemoyer, and
  Hajishirzi]{Min2022MetaICLLT}
Sewon Min, Mike Lewis, Luke Zettlemoyer, and Hannaneh Hajishirzi.
\newblock Metaicl: Learning to learn in context.
\newblock \emph{ArXiv}, abs/2110.15943, 2022{\natexlab{a}}.

\bibitem[Min et~al.(2022{\natexlab{b}})Min, Lyu, Holtzman, Artetxe, Lewis,
  Hajishirzi, and Zettlemoyer]{Min2022RethinkingTR}
Sewon Min, Xinxi Lyu, Ari Holtzman, Mikel Artetxe, Mike Lewis, Hannaneh
  Hajishirzi, and Luke Zettlemoyer.
\newblock Rethinking the role of demonstrations: What makes in-context learning
  work?
\newblock \emph{ArXiv}, abs/2202.12837, 2022{\natexlab{b}}.

\bibitem[Mostafazadeh et~al.(2017)Mostafazadeh, Roth, Louis, Chambers, and
  Allen]{mostafazadeh2017lsdsem}
Nasrin Mostafazadeh, Michael Roth, Annie Louis, Nathanael Chambers, and James
  Allen.
\newblock Lsdsem 2017 shared task: The story cloze test.
\newblock In \emph{Proceedings of the 2nd Workshop on Linking Models of
  Lexical, Sentential and Discourse-level Semantics}, pp.\  46--51, 2017.

\bibitem[Nakano et~al.(2021)Nakano, Hilton, Balaji, Wu, Ouyang, Kim, Hesse,
  Jain, Kosaraju, Saunders, Jiang, Cobbe, Eloundou, Krueger, Button, Knight,
  Chess, and Schulman]{Nakano2021WebGPTBQ}
Reiichiro Nakano, Jacob Hilton, S.~Arun Balaji, Jeff Wu, Long Ouyang, Christina
  Kim, Christopher Hesse, Shantanu Jain, Vineet Kosaraju, William Saunders,
  Xu~Jiang, Karl Cobbe, Tyna Eloundou, Gretchen Krueger, Kevin Button, Matthew
  Knight, Benjamin Chess, and John Schulman.
\newblock Webgpt: Browser-assisted question-answering with human feedback.
\newblock \emph{ArXiv}, abs/2112.09332, 2021.

\bibitem[Ouyang et~al.(2022)Ouyang, Wu, Jiang, Almeida, Wainwright, Mishkin,
  Zhang, Agarwal, Slama, Ray, Schulman, Hilton, Kelton, Miller, Simens, Askell,
  Welinder, Christiano, Leike, and Lowe]{Ouyang2022TrainingLM}
Long Ouyang, Jeff Wu, Xu~Jiang, Diogo Almeida, Carroll~L. Wainwright, Pamela
  Mishkin, Chong Zhang, Sandhini Agarwal, Katarina Slama, Alex Ray, John
  Schulman, Jacob Hilton, Fraser Kelton, Luke~E. Miller, Maddie Simens, Amanda
  Askell, Peter Welinder, Paul~Francis Christiano, Jan Leike, and Ryan~J. Lowe.
\newblock Training language models to follow instructions with human feedback.
\newblock \emph{ArXiv}, abs/2203.02155, 2022.

\bibitem[Perez et~al.(2022)Perez, Huang, Song, Cai, Ring, Aslanides, Glaese,
  McAleese, and Irving]{Perez2022RedTL}
Ethan Perez, Saffron Huang, Francis Song, Trevor Cai, Roman Ring, John
  Aslanides, Amelia Glaese, Nathan McAleese, and Geoffrey Irving.
\newblock Red teaming language models with language models.
\newblock \emph{ArXiv}, abs/2202.03286, 2022.

\bibitem[Raffel et~al.(2020)Raffel, Shazeer, Roberts, Lee, Narang, Matena,
  Zhou, Li, and Liu]{Raffel2020ExploringTL}
Colin Raffel, Noam~M. Shazeer, Adam Roberts, Katherine Lee, Sharan Narang,
  Michael Matena, Yanqi Zhou, Wei Li, and Peter~J. Liu.
\newblock Exploring the limits of transfer learning with a unified text-to-text
  transformer.
\newblock \emph{ArXiv}, abs/1910.10683, 2020.

\bibitem[Rajpurkar et~al.(2016)Rajpurkar, Zhang, Lopyrev, and
  Liang]{rajpurkar2016squad}
Pranav Rajpurkar, Jian Zhang, Konstantin Lopyrev, and Percy Liang.
\newblock Squad: 100,000+ questions for machine comprehension of text.
\newblock \emph{arXiv preprint arXiv:1606.05250}, 2016.

\bibitem[Roemmele et~al.(2011)Roemmele, Bejan, and Gordon]{roemmele2011choice}
Melissa Roemmele, Cosmin~Adrian Bejan, and Andrew~S Gordon.
\newblock Choice of plausible alternatives: An evaluation of commonsense causal
  reasoning.
\newblock In \emph{2011 AAAI Spring Symposium Series}, 2011.

\bibitem[Roller et~al.(2021)Roller, Dinan, Goyal, Ju, Williamson, Liu, Xu, Ott,
  Shuster, Smith, Boureau, and Weston]{Roller2021RecipesFB}
Stephen Roller, Emily Dinan, Naman Goyal, Da~Ju, Mary Williamson, Yinhan Liu,
  Jing Xu, Myle Ott, Kurt Shuster, Eric~Michael Smith, Y.-Lan Boureau, and
  Jason Weston.
\newblock Recipes for building an open-domain chatbot.
\newblock In \emph{EACL}, 2021.

\bibitem[Sanh et~al.(2021)Sanh, Webson, Raffel, Bach, Sutawika, Alyafeai,
  Chaffin, Stiegler, Scao, Raja, Dey, BARI, Xu, Thakker, Sharma, Szczechla,
  Kim, Chhablani, Nayak, Datta, Chang, Jiang, Wang, Manica, Shen, Yong, Pandey,
  Bawden, Wang, Neeraj, Rozen, Sharma, Santilli, F{\'e}vry, Fries, Teehan,
  Biderman, Gao, Bers, Wolf, and Rush]{Sanh2021MultitaskPT}
Victor Sanh, Albert Webson, Colin Raffel, Stephen~H. Bach, Lintang~A. Sutawika,
  Zaid Alyafeai, Antoine Chaffin, Arnaud Stiegler, Teven~Le Scao, Arun Raja,
  Manan Dey, M~SAIFUL BARI, Canwen Xu, Urmish Thakker, Shanya Sharma, Eliza
  Szczechla, Taewoon Kim, Gunjan Chhablani, Nihal~V. Nayak, Debajyoti Datta,
  Jonathan Chang, Mike Tian-Jian Jiang, Han Wang, Matteo Manica, Sheng Shen,
  Zheng~Xin Yong, Harshit Pandey, Rachel Bawden, Thomas Wang, Trishala Neeraj,
  Jos Rozen, Abheesht Sharma, Andrea Santilli, Thibault F{\'e}vry, Jason~Alan
  Fries, Ryan Teehan, Stella~Rose Biderman, Leo Gao, T.~G.~Owe Bers, Thomas
  Wolf, and Alexander~M. Rush.
\newblock Multitask prompted training enables zero-shot task generalization.
\newblock \emph{ArXiv}, abs/2110.08207, 2021.

\bibitem[Stiennon et~al.(2020)Stiennon, Ouyang, Wu, Ziegler, Lowe, Voss,
  Radford, Amodei, and Christiano]{Stiennon2020LearningTS}
Nisan Stiennon, Long Ouyang, Jeff Wu, Daniel~M. Ziegler, Ryan~J. Lowe, Chelsea
  Voss, Alec Radford, Dario Amodei, and Paul Christiano.
\newblock Learning to summarize from human feedback.
\newblock \emph{ArXiv}, abs/2009.01325, 2020.

\bibitem[Thorne et~al.(2018)Thorne, Vlachos, Christodoulopoulos, and
  Mittal]{Thorne2018FEVERAL}
James Thorne, Andreas Vlachos, Christos Christodoulopoulos, and Arpit Mittal.
\newblock Fever: a large-scale dataset for fact extraction and verification.
\newblock \emph{ArXiv}, abs/1803.05355, 2018.

\bibitem[Wang et~al.(2018)Wang, Singh, Michael, Hill, Levy, and
  Bowman]{wang2018glue}
Alex Wang, Amanpreet Singh, Julian Michael, Felix Hill, Omer Levy, and Samuel~R
  Bowman.
\newblock Glue: A multi-task benchmark and analysis platform for natural
  language understanding.
\newblock \emph{arXiv preprint arXiv:1804.07461}, 2018.

\bibitem[Wang \& Komatsuzaki(2021)Wang and Komatsuzaki]{gpt-j}
Ben Wang and Aran Komatsuzaki.
\newblock {GPT-J-6B: A 6 Billion Parameter Autoregressive Language Model}.
\newblock \url{https://github.com/kingoflolz/mesh-transformer-jax}, May 2021.

\bibitem[Wei et~al.(2022{\natexlab{a}})Wei, Bosma, Zhao, Guu, Yu, Lester, Du,
  Dai, and Le]{Wei2022FinetunedLM}
Jason Wei, Maarten Bosma, Vincent Zhao, Kelvin Guu, Adams~Wei Yu, Brian Lester,
  Nan Du, Andrew~M. Dai, and Quoc~V. Le.
\newblock Finetuned language models are zero-shot learners.
\newblock \emph{ArXiv}, abs/2109.01652, 2022{\natexlab{a}}.

\bibitem[Wei et~al.(2022{\natexlab{b}})Wei, Wang, Schuurmans, Bosma, Chi, Le,
  and Zhou]{Wei2022ChainOT}
Jason Wei, Xuezhi Wang, Dale Schuurmans, Maarten Bosma, Ed~Chi, Quoc Le, and
  Denny Zhou.
\newblock Chain of thought prompting elicits reasoning in large language
  models.
\newblock \emph{ArXiv}, abs/2201.11903, 2022{\natexlab{b}}.

\bibitem[Weidinger et~al.(2021)Weidinger, Mellor, Rauh, Griffin, Uesato, Huang,
  Cheng, Glaese, Balle, Kasirzadeh, Kenton, Brown, Hawkins, Stepleton, Biles,
  Birhane, Haas, Rimell, Hendricks, Isaac, Legassick, Irving, and
  Gabriel]{Weidinger2021EthicalAS}
Laura Weidinger, John F.~J. Mellor, Maribeth Rauh, Conor Griffin, Jonathan
  Uesato, Po-Sen Huang, Myra Cheng, Mia Glaese, Borja Balle, Atoosa Kasirzadeh,
  Zachary Kenton, Sande~Minnich Brown, William~T. Hawkins, Tom Stepleton,
  Courtney Biles, Abeba Birhane, Julia Haas, Laura Rimell, Lisa~Anne Hendricks,
  William~S. Isaac, Sean Legassick, Geoffrey Irving, and Iason Gabriel.
\newblock Ethical and social risks of harm from language models.
\newblock \emph{ArXiv}, abs/2112.04359, 2021.

\bibitem[Wolf et~al.(2019)Wolf, Debut, Sanh, Chaumond, Delangue, Moi, Cistac,
  Rault, Louf, Funtowicz, and Brew]{Wolf2019HuggingFacesTS}
Thomas Wolf, Lysandre Debut, Victor Sanh, Julien Chaumond, Clement Delangue,
  Anthony Moi, Pierric Cistac, Tim Rault, R{\'e}mi Louf, Morgan Funtowicz, and
  Jamie Brew.
\newblock Huggingface's transformers: State-of-the-art natural language
  processing.
\newblock \emph{ArXiv}, abs/1910.03771, 2019.

\bibitem[Yin et~al.(2020)Yin, Rajani, Radev, Socher, and
  Xiong]{Yin2020UniversalNL}
Wenpeng Yin, Nazneen Rajani, Dragomir Radev, Richard Socher, and Caiming Xiong.
\newblock Universal natural language processing with limited annotations: Try
  few-shot textual entailment as a start.
\newblock In \emph{EMNLP}, 2020.

\bibitem[Zhang et~al.(2015)Zhang, Zhao, and LeCun]{zhang2015character}
Xiang Zhang, Junbo Zhao, and Yann LeCun.
\newblock Character-level convolutional networks for text classification.
\newblock \emph{Advances in neural information processing systems},
  28:\penalty0 649--657, 2015.

\bibitem[Zhao et~al.(2021)Zhao, Wallace, Feng, Klein, and
  Singh]{Zhao2021CalibrateBU}
Tony Zhao, Eric Wallace, Shi Feng, Dan Klein, and Sameer Singh.
\newblock Calibrate before use: Improving few-shot performance of language
  models.
\newblock \emph{ArXiv}, abs/2102.09690, 2021.

\bibitem[Zhong et~al.(2021)Zhong, Lee, Zhang, and Klein]{Zhong2021AdaptingLM}
Ruiqi Zhong, Kristy Lee, Zheng Zhang, and Dan Klein.
\newblock Adapting language models for zero-shot learning by meta-tuning on
  dataset and prompt collections.
\newblock In \emph{EMNLP}, 2021.

\bibitem[Zhou et~al.(2022)Zhou, He, Ma, Berg-Kirkpatrick, and
  Neubig]{Zhou2022PromptCF}
Chunting Zhou, Junxian He, Xuezhe Ma, Taylor Berg-Kirkpatrick, and Graham
  Neubig.
\newblock Prompt consistency for zero-shot task generalization.
\newblock \emph{ArXiv}, abs/2205.00049, 2022.

\end{thebibliography}
